\documentclass[final]{cvpr}

\usepackage{times}
\usepackage{epsfig}
\usepackage{listings}
\usepackage{graphicx}
\usepackage{amsmath}
\usepackage{amssymb}
\usepackage{hhline}
\usepackage{subcaption}
\usepackage{multirow}
\usepackage[subtle]{savetrees}

\usepackage[pagebackref=true,breaklinks=true,colorlinks,bookmarks=false]{hyperref}

\def\cvprPaperID{****} 
\def\confYear{CVPR 2021}

\begin{document}

\title{The 2021 Hotel-ID to Combat Human Trafficking Competition Dataset}

\author{Rashmi Kamath$^1$, Gregory Rolwes$^1$, Samuel Black$^2$ and Abby Stylianou$^1$\\
Saint Louis University$^1$, Temple University$^2$\\
Corresponding author: {\tt\small rashmi.kamath@slu.edu}
}

\maketitle

\begin{abstract}
   Hotel recognition is an important task for human trafficking investigations since victims are often photographed in hotel rooms. Identifying these hotels is vital to trafficking investigations since they can help track down current and future victims who might be taken to the same places. Hotel recognition is a challenging fine grained visual classification task as there can be little similarity between different rooms within the same hotel, and high similarity between rooms from different hotels (especially if they are from the same chain). Hotel recognition to combat human trafficking poses additional challenges as investigative images are often low quality, contain uncommon camera angles and are highly occluded. Here, we present the 2021 Hotel-ID dataset to help raise awareness for this problem and generate novel approaches. The dataset consists of hotel room images that have been crowd-sourced and uploaded through the TraffickCam mobile application. The quality of these images is similar to investigative images and hence models trained on these images have good chances of accurately narrowing down on the correct hotel.
\end{abstract}

\section{Introduction}
\begin{figure}
    \centering
    \includegraphics[width=\columnwidth]{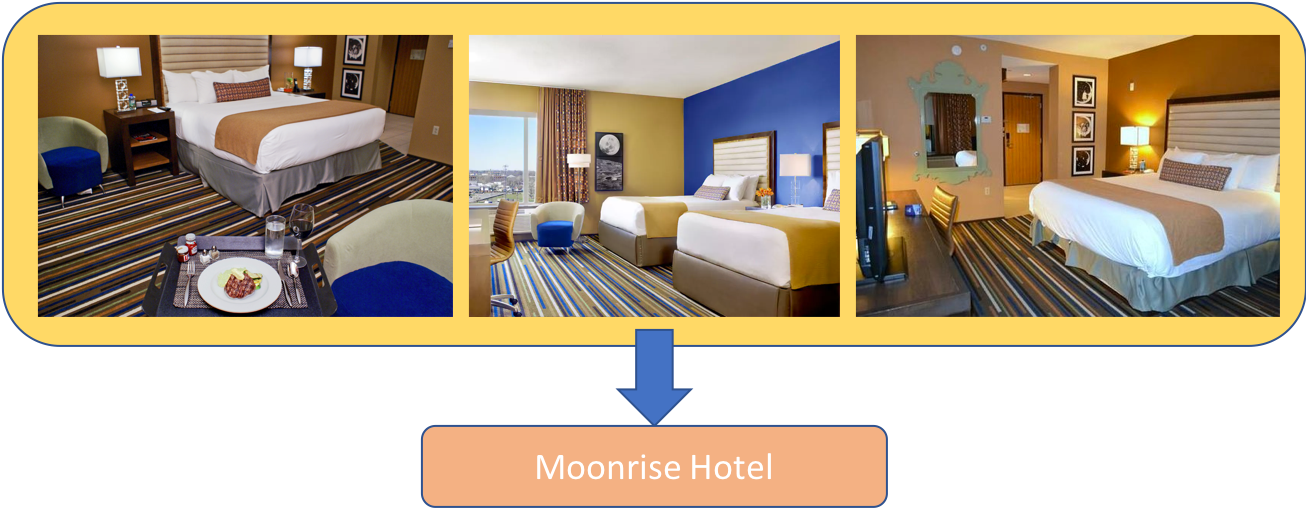}
    \caption[Hotel recognition]{Hotel recognition is the task of identifying what hotel is seen in an image captured in a hotel room.}
    \label{fig:hotelRecognition}
\end{figure}

Hotel recognition is the task of identifying the hotel in images taken in a hotel room, as seen in Figure~\ref{fig:hotelRecognition}. At first glance, this problem may seem easier than other scene recognition problems -- hotels contain a relatively limited set of different objects compared to the space of all possible outdoor scenes, and one room within a hotel is likely visually quite similar to another room. These properties, however, turn out to be both challenging, and not always correct.

\begin{figure}
    \centering
    \includegraphics[width=\columnwidth]{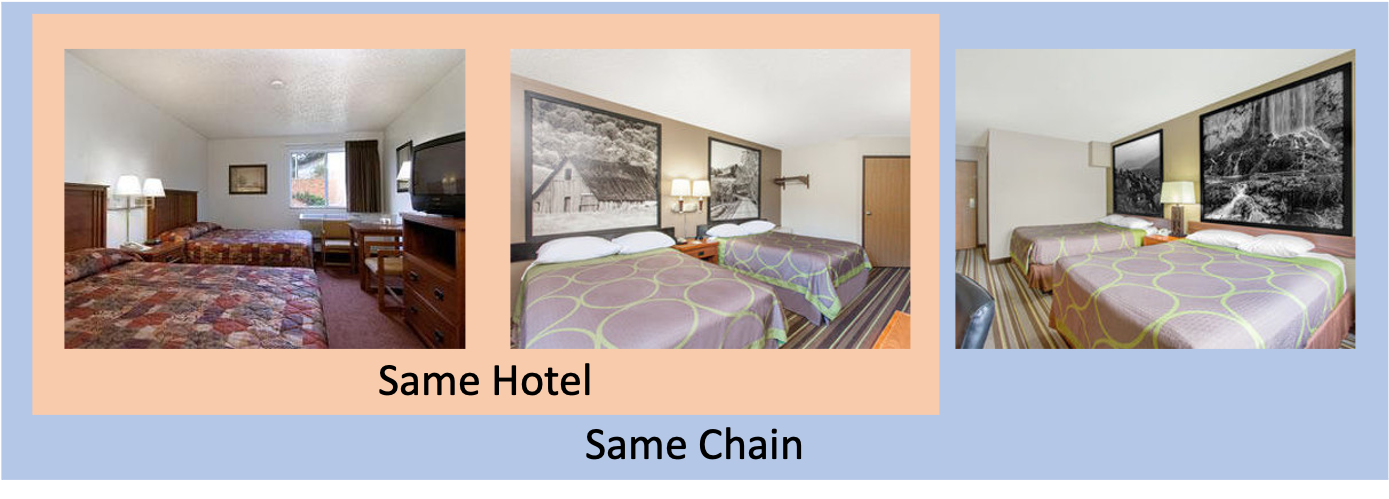}
    \caption[Visual similarity in hotel instances and hotel chains]{One of the main challenges with hotel recognition is that images within the same hotel, such as the left and center images in this figure, may be visually dissimilar, while images from different hotels, especially those from the same chain, such as the center and right images, may be visually similar.}
    \label{fig:sameHotel_vs_sameChain2}
\end{figure}

While there may be a limited number of objects in a hotel rooms (e.g., a headboard, lamps, artwork, etc.), these objects are often found in a wide variety of different configurations.  In contrast, even within the challenging domain of scene localization, the real world features typically have a fixed 3D configuration that can be (implicitly or explicitly) used to understand if they are consistent with a 2D image.

Not only is the spatial configuration of hotel rooms within a hotel complex, it is often the case that different rooms within a hotel may have significantly different visual appearances, with limited or non-existent overlap between the objects in the rooms.  Further complicating the situation, two rooms in \textit{different} hotels, but from the same hotel \textit{chain}, may be more visually similar than two rooms from the same hotel. This can be seen in Figure~\ref{fig:sameHotel_vs_sameChain2}, where the first two images are from the same hotel before and after a renovation, and the second two images are from two different hotels in the same chain. The only overlapping feature between the two images from the same hotel is the wall-mounted lamp, while almost every other feature is more similar to the image from a different hotel in the same chain.

Hotel recognition, the ability to recognize a hotel from an image taken in a hotel room, is therefore a challenging fine grained visual categorization task. Additionally, it has a specific societally important use-case related to human trafficking. In recent years, the number of images of victims of human trafficking shared online has grown at an alarming rate~\cite{bouche2015report,ncmecAmicusBrief}. Whether used for advertising or exchanged among criminal networks, these photographs serve as visual evidence of where the victim was trafficked. Such images are often captured in hotel rooms. Identifying the hotels in these photographs gives insight into where a trafficking victim has been moved previously and where their trafficker may move them or others in the future. Understanding how traffickers operate, and how to rescue victims is a top priority for law enforcement~\cite{nationalStrategy}.

In this paper, we present the 2021 Hotel-ID dataset and competition to raise awareness for this problem and generate novel approaches to help address it. The dataset consists of images uploaded to the TraffickCam mobile application~\cite{aipr2015,aipr2017}, which allows every day travellers to upload images of their hotel rooms specifically with the goal of helping to combat human trafficking. The entire TraffickCam dataset, along with publicly available travel images, are then indexed and made available through an investigative search system at the National Center for Missing and Exploited Children~\cite{tcam2019}. In Section~\ref{sec:ethical}, we discuss the ethical concerns of building a tool that can identify locations in imagery, making those tools available to investigators, and how we've worked to mitigate those concerns both in the broader system and within this dataset and competition.

\section{Dataset}
The dataset for this challenge overlaps with the much larger Hotels-50K dataset~\cite{hotels50k}, which included over a million images from 50,000 hotels. This dataset is intended to be more approachable for researchers, while still large enough to highlight the challenges of hotel recognition at scale. The 2021 Hotel-ID dataset consists of a training set with 97,527 images from 7,770 hotels across the globe. These hotels are additionally categorized by which of 86 different hotel chains they belong to, if known.

\begin{figure}
    \centering
    \includegraphics[width=\columnwidth]{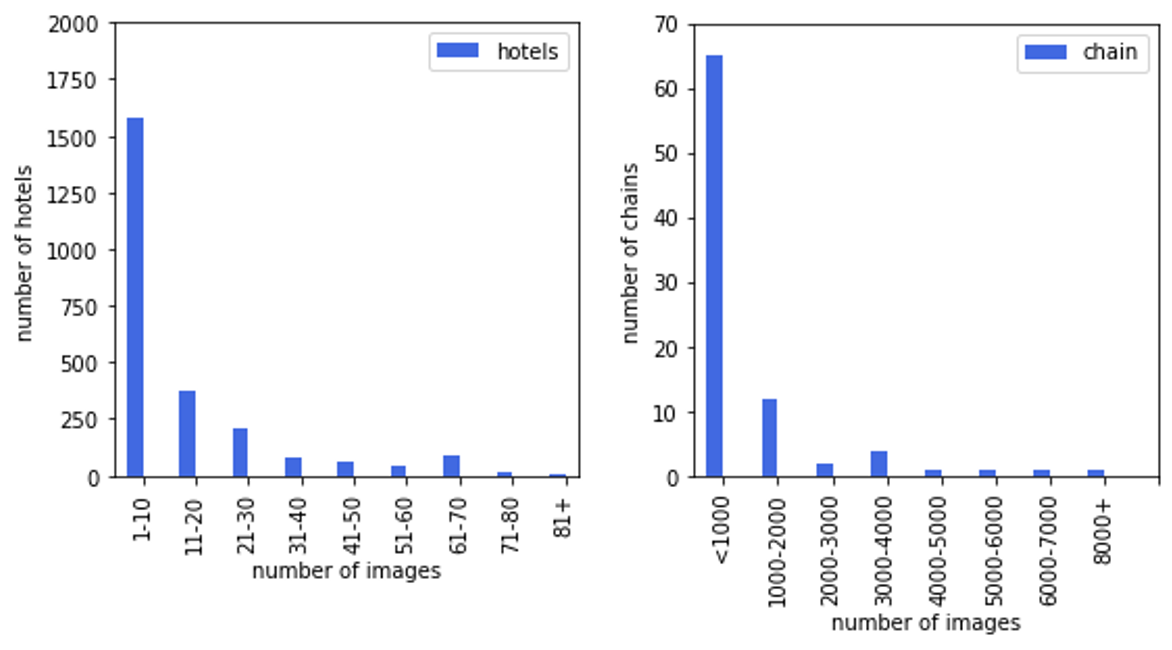}
    \caption{Image distribution by hotels (left) and chains (right) in the dataset.}
    \label{fig:Hotels}
\end{figure}

Figure~\ref{fig:Hotels} shows the histogram of the number of images per hotel and per chain within the training dataset. Some hotels, and some chains, are more prevalent in the dataset than others. This is due both to actual differences in prevalence of certain hotels and chains in the world, as well as biases in where TraffickCam images are captured (e.g., travelers submit more pictures from tourist destinations than from side-of-the-road motels).


\section{Evaluation}
The test dataset consists of 12,400 images, also from the TraffickCam application. Each image in the test set was taken in a hotel that is represented in the training set. No additional meta-data is provided for the test images.
The test set images were selected to ensure that they have been captured by a different user than the ones in the training set images, as re-identifying the same exact room photographed by the same user on the same device is less challenging than the real-world recognition challenge. The evaluation task is to accurately identify the correct Hotel instance for the images in the test set.

\subsection{Evaluation Metrics}
The Hotel identification task is a classification task in which the results should be provided as the five most probable hotel IDs for an image in the test set. The evaluation metrics is the Mean Average Precision at 5 (MAP@5) score:

\begin{align*}
 MAP@5&=\frac{1}{U}\sum_{u=1}^{U}\sum_{k=1}^{min(n,5)}P(k) \times \text{rel}(k)
\end{align*}

where U is the number of images, P(k) is the precision at cutoff k, n is the number of predictions per image, and rel(k) is an indicator function equaling 1 if the item at rank k is a relevant correct label, zero otherwise.

We additionally report on image retrieval performance, using recall at 1, 10 and 100. This answers the question ``was an image from the same class found in the top K results?''. While the classification metric is well suited for the 2021 Hotel-ID dataset and competition, the real world use case involves attempting to identify which of over half a million hotels world wide is seen in a query image. Classification problems with this large a number of classes are often approached as image retrieval problems, where the class is inferred using, for example, k-Nearest Neighbors.

\section{Baseline Approaches}
To set the Baseline for this challenge, we trained two different types of approaches: first, a classification network trained with cross-entropy loss, and second, a series of Deep Metric Learning that have previously been used for hotel recognition (Batch-All~\cite{song2016deep}, Easy Positive Hard Negative (EPHN)~\cite{epshn}). We additionally compare against a pre-trained model trained on the prior Hotels-50K dataset~\cite{hotels50k} using Selectively Contrastive Triplet Loss (SCT)~\cite{hard_negatives}, the current state of the art approach on that dataset.

The model used in all cases was a Resnet-50 model~\cite{resnet}, pre-trained on ImageNet~\cite{deng2009imagenet}. For the classification approach, the network has a 8,000 dimensional output feature; for the metric learning approaches, we output 256-dimensional feature embeddings. We train with images that are resized to 256x256 and randomly cropped to 224x224 (at test time we use center crops). All approaches are trained with 128 image batches. Images within a batch are additionally randomly flipped horizontally, rotated between -30 to +30 degrees, and random color jittering is performed, following the data augmentation from the Hotels-50K paper. For the classification-based approach, we perform 50\% dropout on the weights of the final fully connected layer and use label smoothing (0.1)~\cite{label_smoothing} in order to effectively train a classification network with so many classes.

For the classification approach, we use the SGD optimizer with a learning rate of 0.2 and cosine annealing~\cite{LoshchilovH16a}, and weight decay of $5\times10^{-4}$. For both Batch-All and EPHN, we use Adam with a learning rate of $1\times10^{-5}$.

For all networks, we report Recall at 1, 10 and 100, as well as MAP@5. In order to compute recall results for the classification network, we compute cosine similarity on the $2048D$ output of the global average pooling layer (MAP@5 is computed using the $8000D$ classification output). In order to compute classification results for the metric learning networks, we compute the MAP@5 from the top 5 images from the 100 nearest neighbors computed based on the cosine similarity of the $256D$ embedding output.

\section{Results}
The comparative results for all three approaches is shown in Table~\ref{tab:Table1}. The mean average precision scores are in line with the retrieval for k=1 scores for each method. The classification-based approach significantly outperforms any of the metric learning approaches. This approach, however, has limitations in the real-world setting, where there are hundreds of thousands of classes of hotels; it is not feasible to train a fully connected layer of that size. Even if that practical limitation did not exist, in the real world setting, we need to be able to perform retrieval on new hotels that may have been added to the dataset without necessarily having to retrain every time a new hotel is introduced. In these cases, it is still practical to treat the hotel recognition problem as an image retrieval problem and use metric learning based approaches.

We additionally show qualitative results in Figure~\ref{fig:example_results}, where we show three example query images from the test set and their five nearest neighbors using each of the different approaches. Results from the correct hotel are outlined in green. The first query seems relatively easy, with each of the different approaches finding results from the correct hotel. The second query is matched by the Selectively Contrastive approach based seemingly on the carpet. In the final query, no correct matches are found by any of the approaches in the top 5 results, but the Cross-Entropy model finds several rooms with extremely similar shelving units, including the first result which is a nearly identical room from a different hotel in the same hotel chain.

\begin{table}
  \centering
\begin{tabular}{ |p{2.2cm}|p{.8cm}|p{.8cm}|p{1cm}||p{1.2cm}|  }
\hline
\textbf{Method} & \textbf{R@1} & \textbf{R@10} & \textbf{R@100} & \textbf{MAP@5} \\
\hline
Cross-Entropy & 49.3 & 68.8 & 84.8 & 55.1 \\
\hhline{|=|=|=|=||=|}
Batch-All~\cite{song2016deep} & 12.17 & 26.49 & 46.54 & 15.45 \\
EPHN~\cite{epshn} & 25.43 & 41.04 & 56.38 & 29.35 \\
SCT~\cite{hard_negatives} & 36.04 & 47.34 & 59.69 & 39.54 \\

\hline
\end{tabular}
\newline
\caption{Results for each of the different baseline approaches. We include Retrieval@K accuracy, with K=1, 10 and 100, as well as Mean Average Precision at 5. The classification approach outperforms any of the metric learning approaches by a significant margin, but the metric learning approaches are more likely to generalize to the real-world domain where there are hundreds of thousands of hotel classes.}
\label{tab:Table1}
\end{table}


\newcommand\imheight{.6in}
\setlength{\tabcolsep}{1pt}
\renewcommand{\arraystretch}{1.5}
\begin{figure*}
\begin{tabular}{|c|c|ccccc|}
\hline
&&\multicolumn{5}{|c|}{Image Retrieval Results}\\
Query Image & Approach & k=1 & 2 & 3 & 4 & 5 \\
\hline
  \multirow{4}{*}{\vspace{-1in}\includegraphics[height=1in]{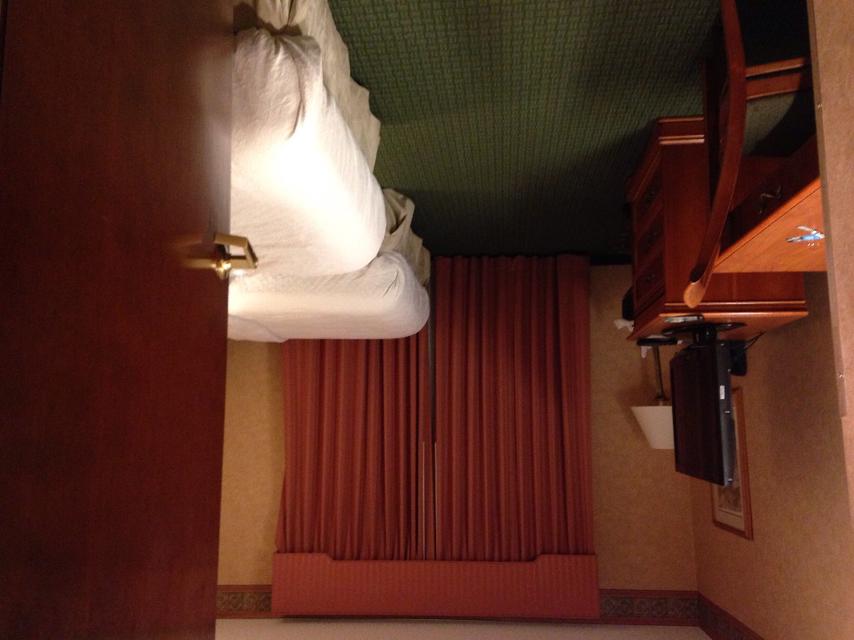}} &
  \raisebox{2\height}{Cross-Entropy} & \includegraphics[height=\imheight]{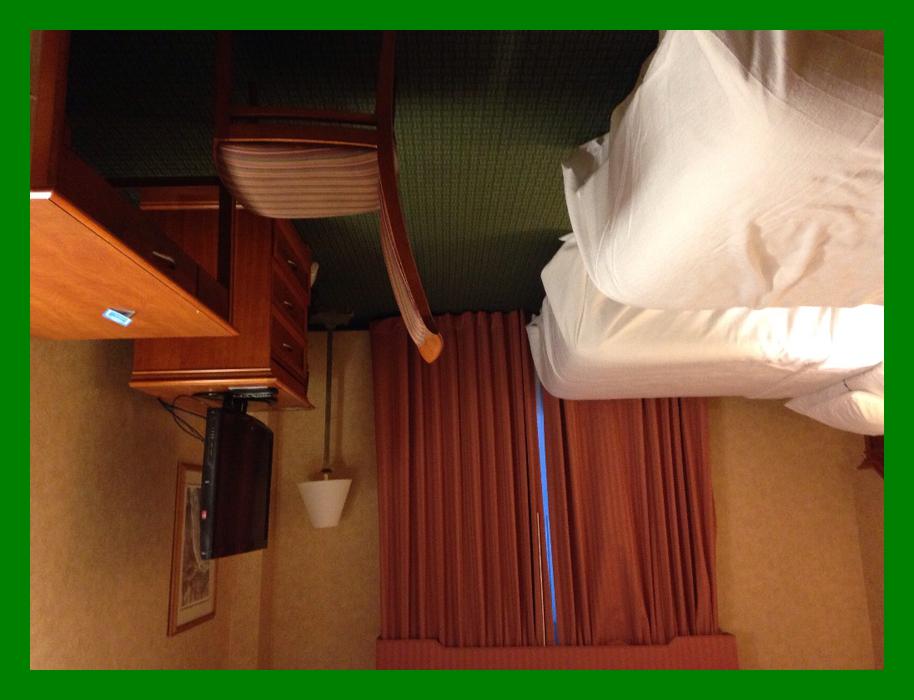} & \includegraphics[height=\imheight]{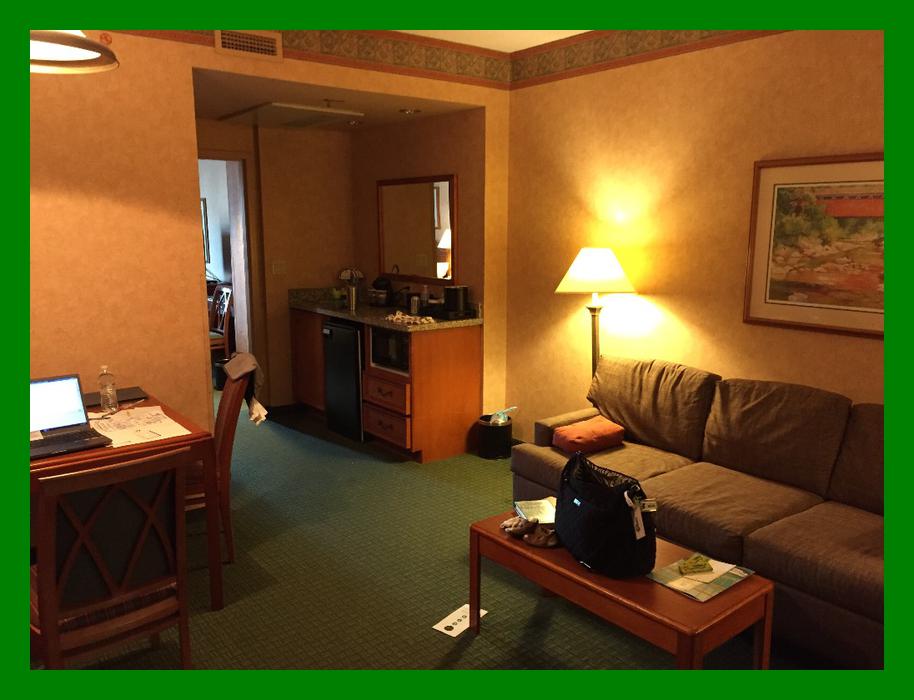} & \includegraphics[height=\imheight]{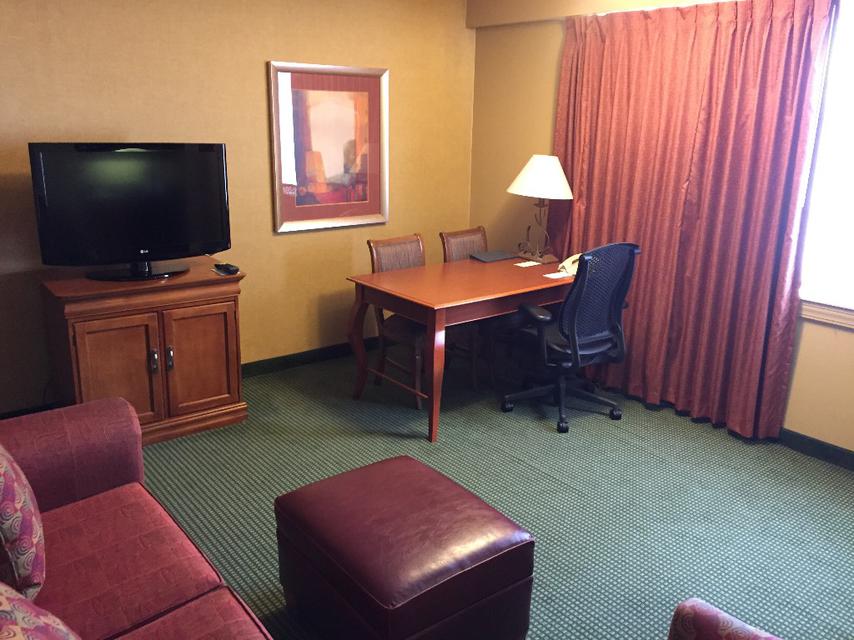} & \includegraphics[height=\imheight]{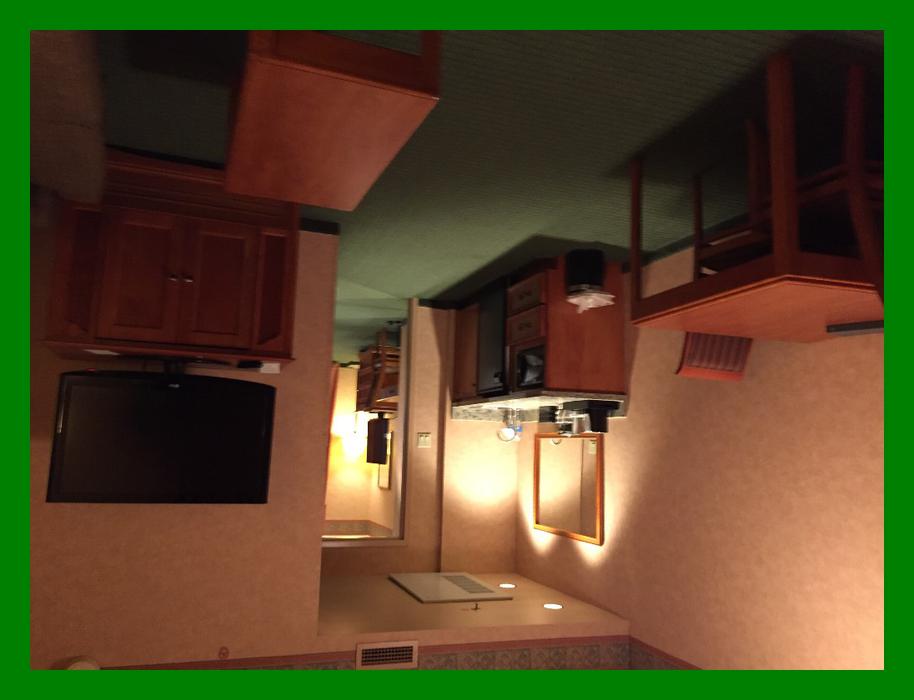} & \includegraphics[height=\imheight]{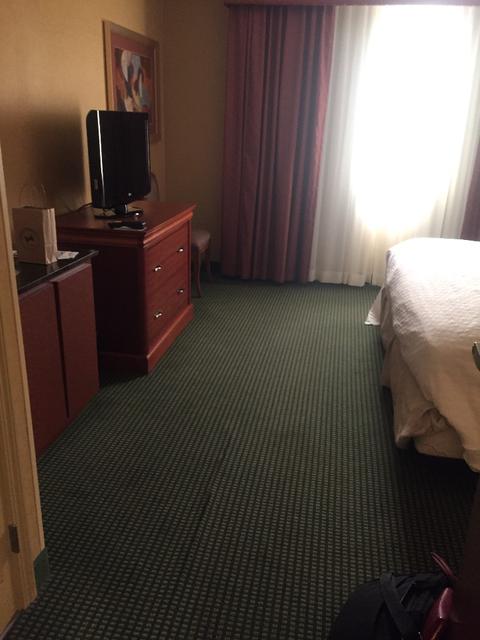} \\
& \raisebox{2\height}{Batch-All} & \includegraphics[height=\imheight]{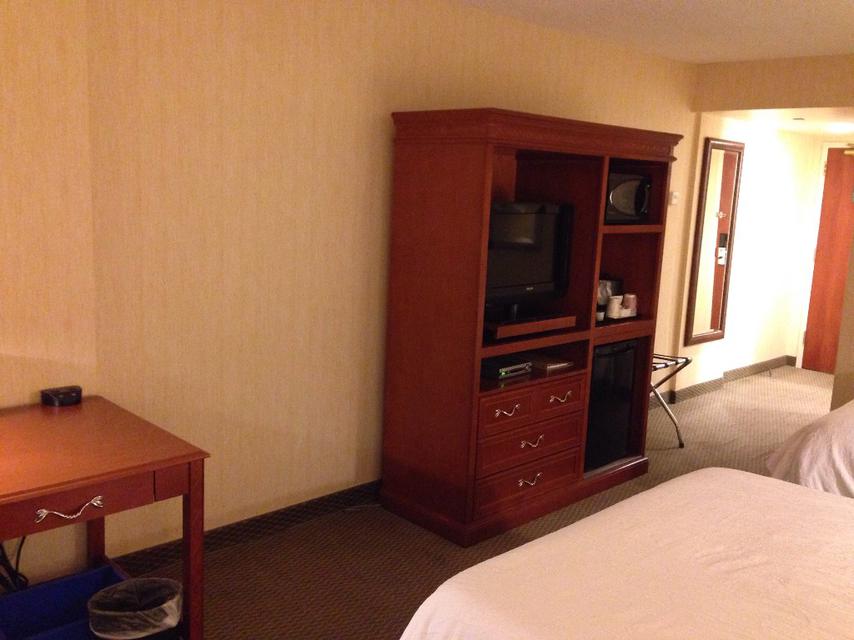} & \includegraphics[height=\imheight]{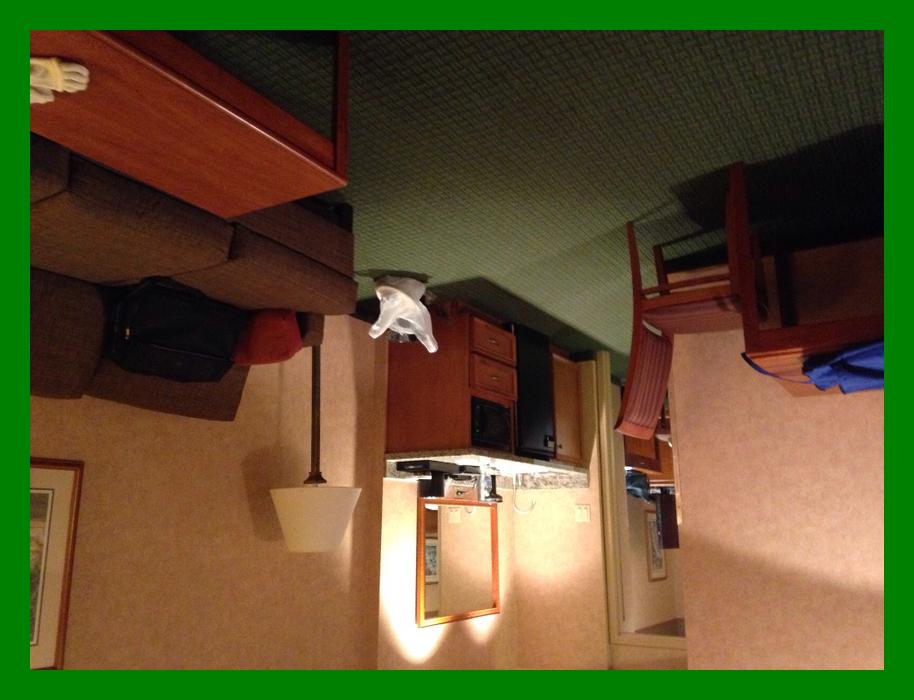} & \includegraphics[height=\imheight]{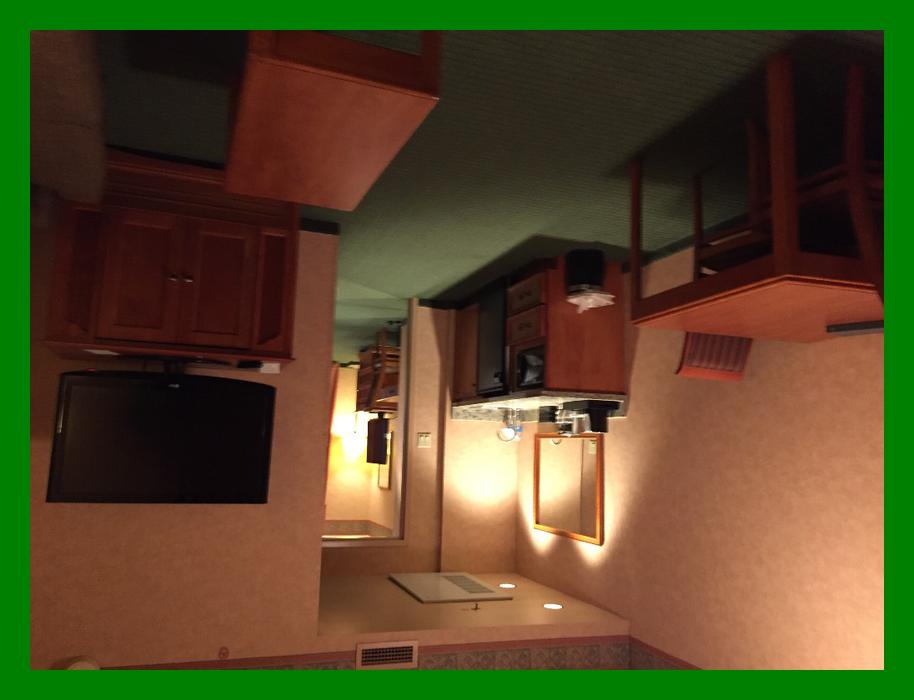} & \includegraphics[height=\imheight]{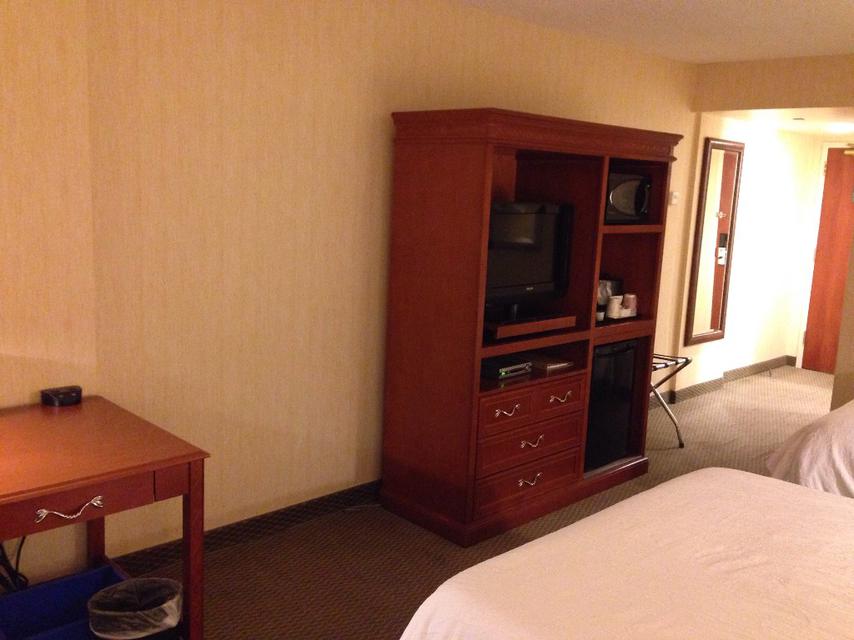} & \includegraphics[height=\imheight]{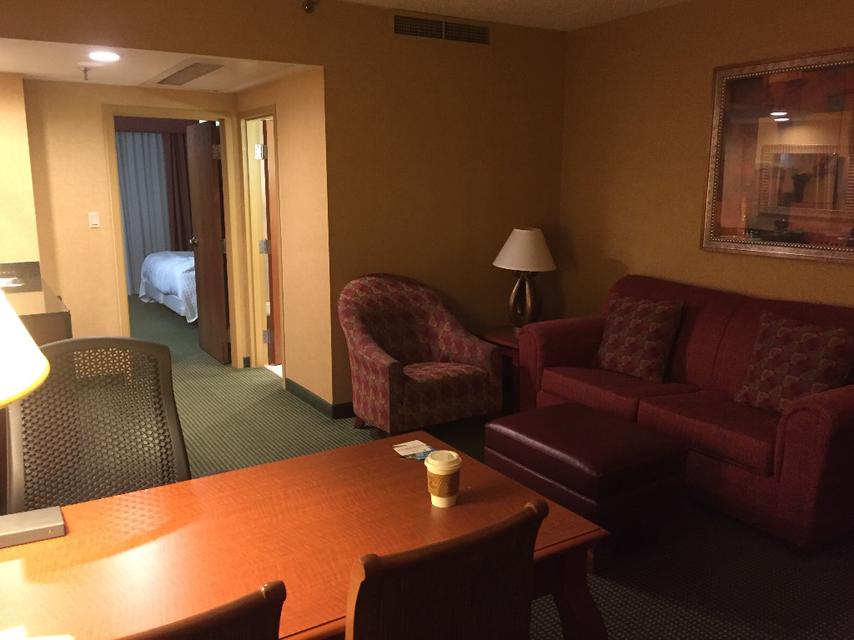} \\
& \raisebox{2\height}{EPHN} & \includegraphics[height=\imheight]{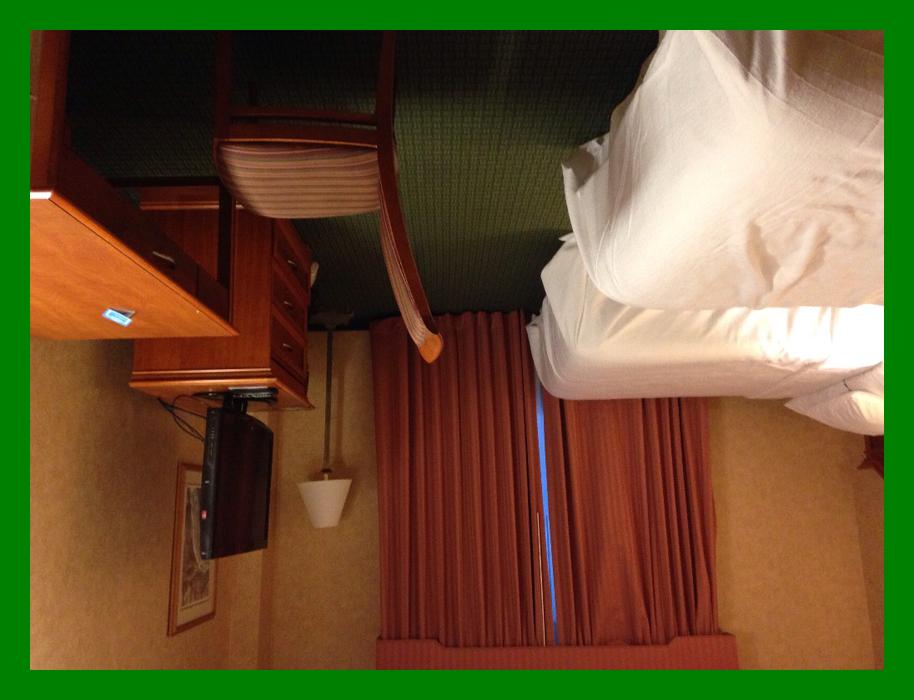} & \includegraphics[height=\imheight]{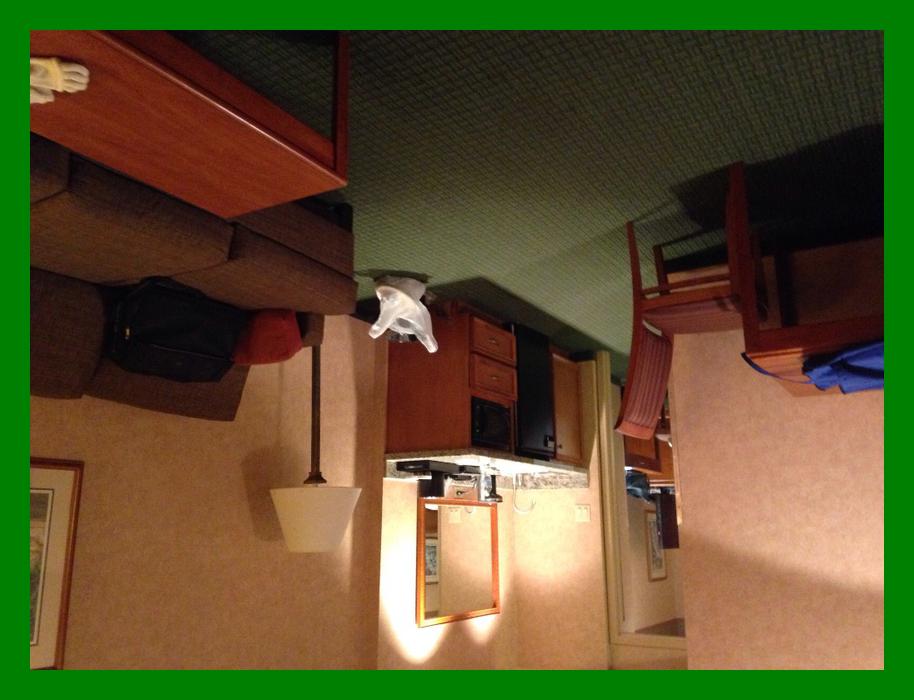} & \includegraphics[height=\imheight]{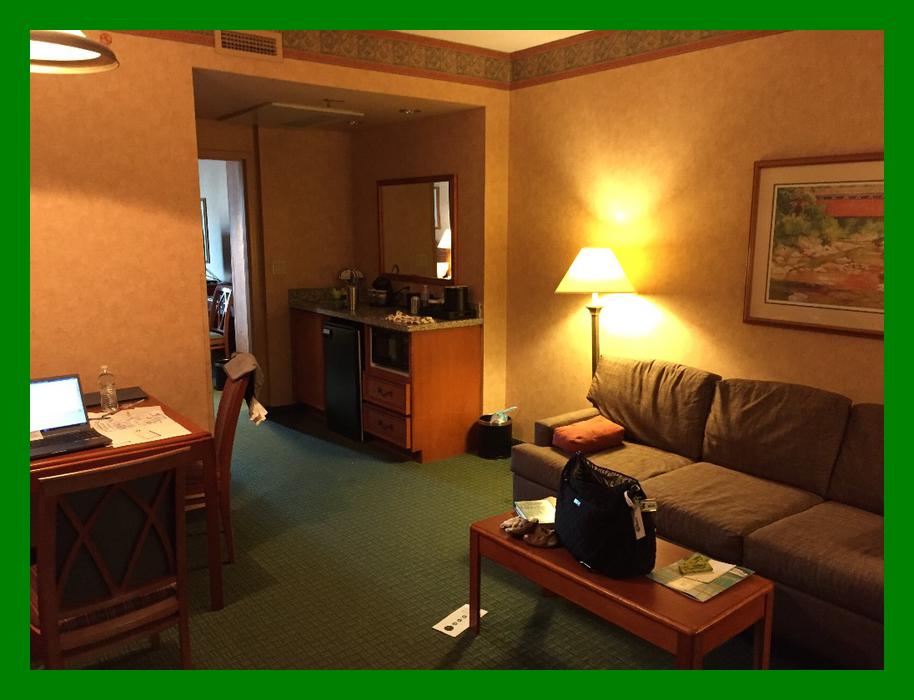} & \includegraphics[height=\imheight]{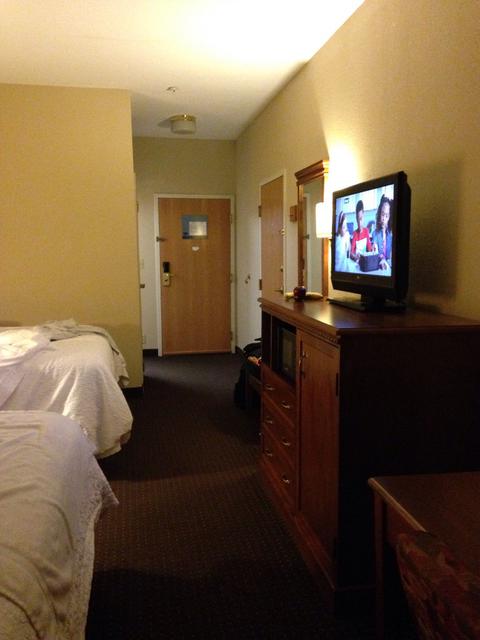} & \includegraphics[height=\imheight]{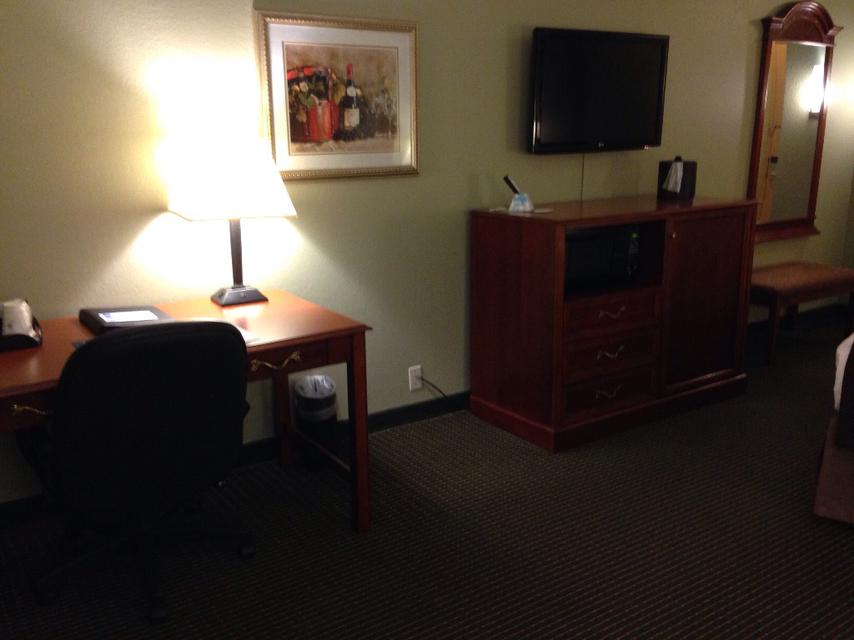} \\
& \raisebox{2\height}{SCT} & \includegraphics[height=\imheight]{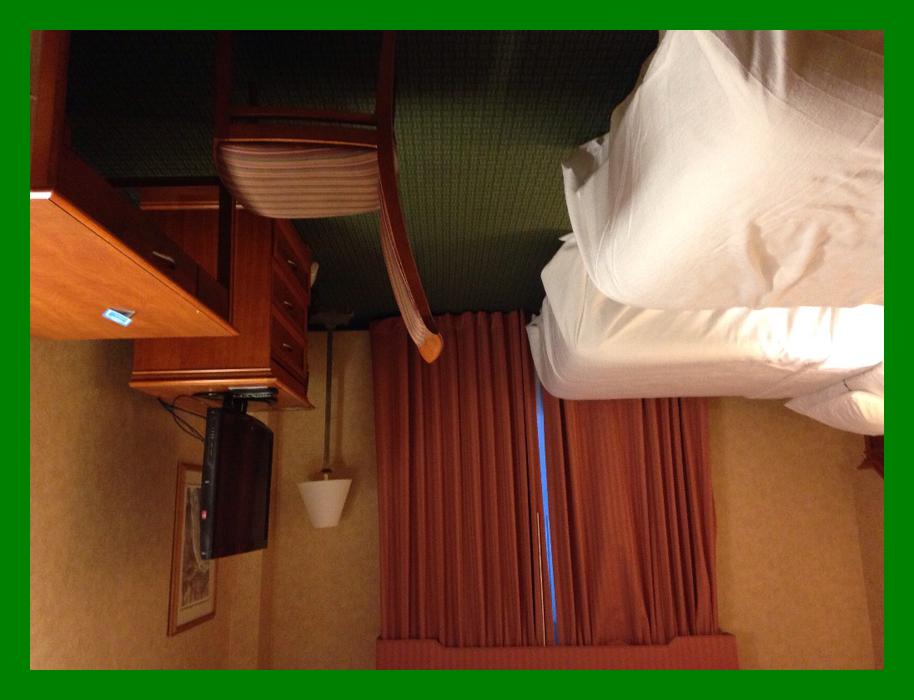} & \includegraphics[height=\imheight]{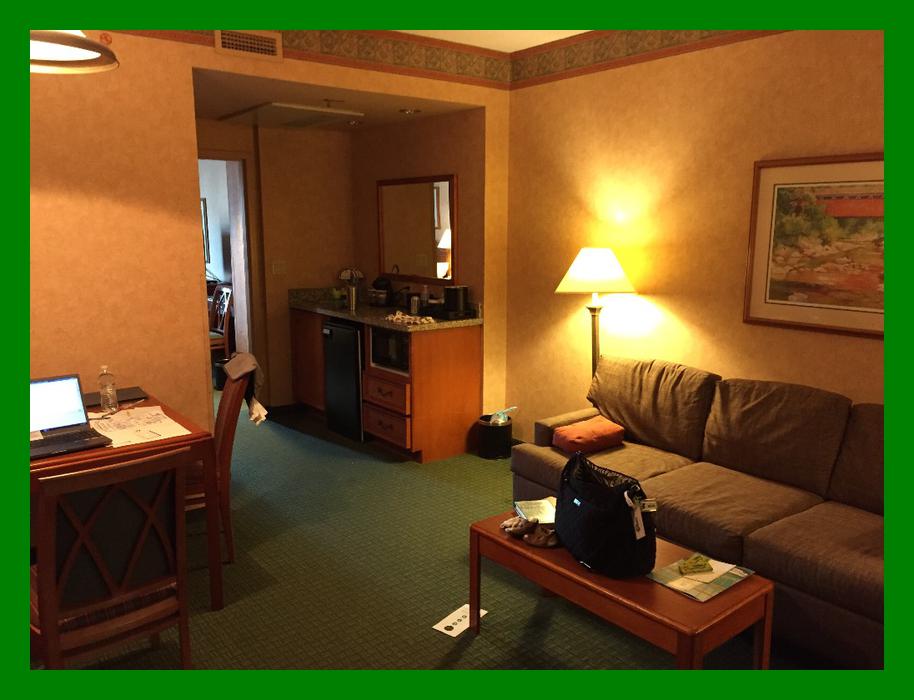} & \includegraphics[height=\imheight]{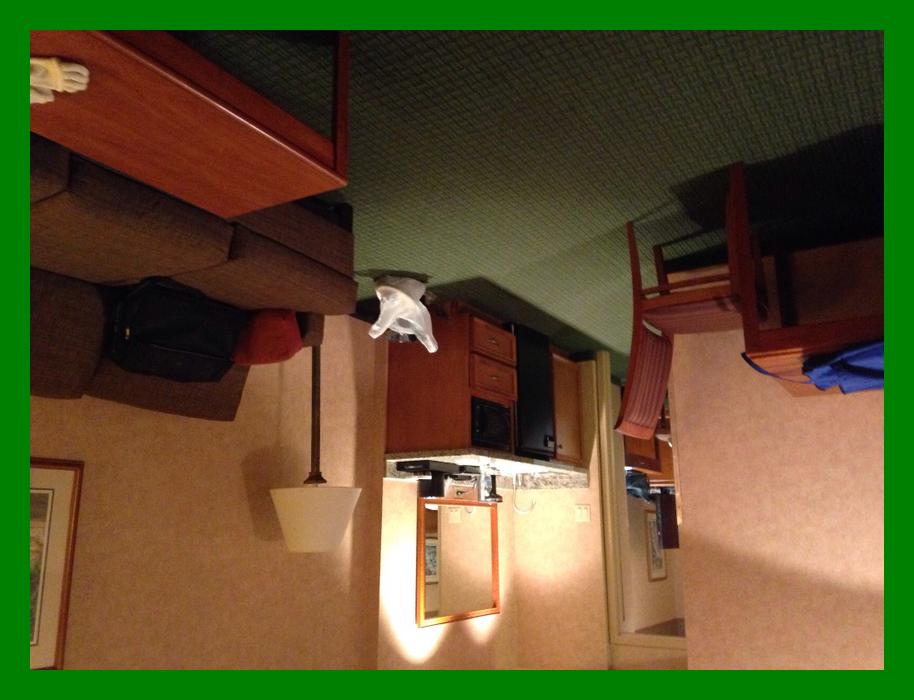} & \includegraphics[height=\imheight]{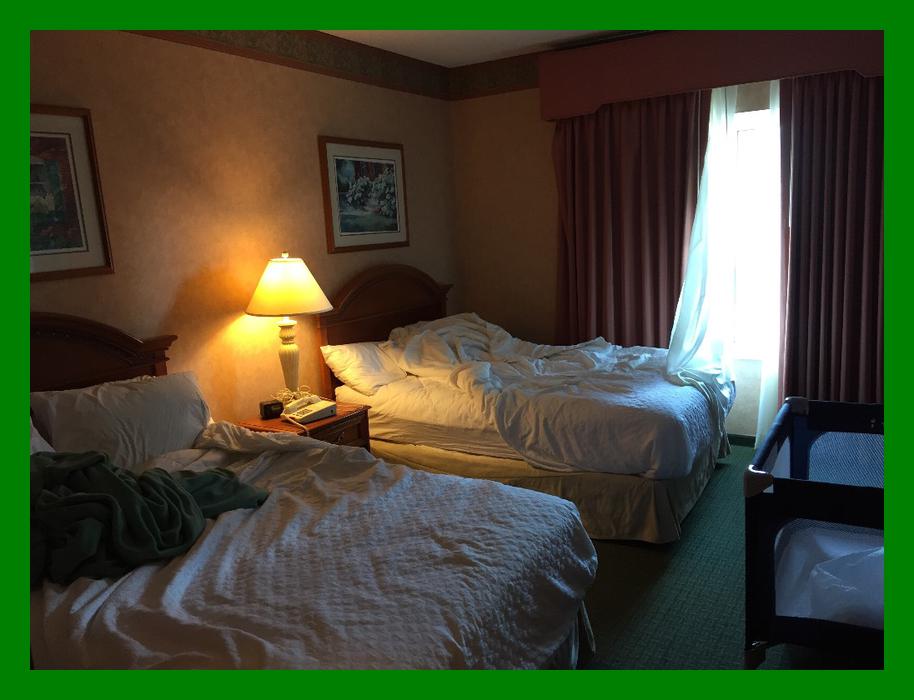} & \includegraphics[height=\imheight]{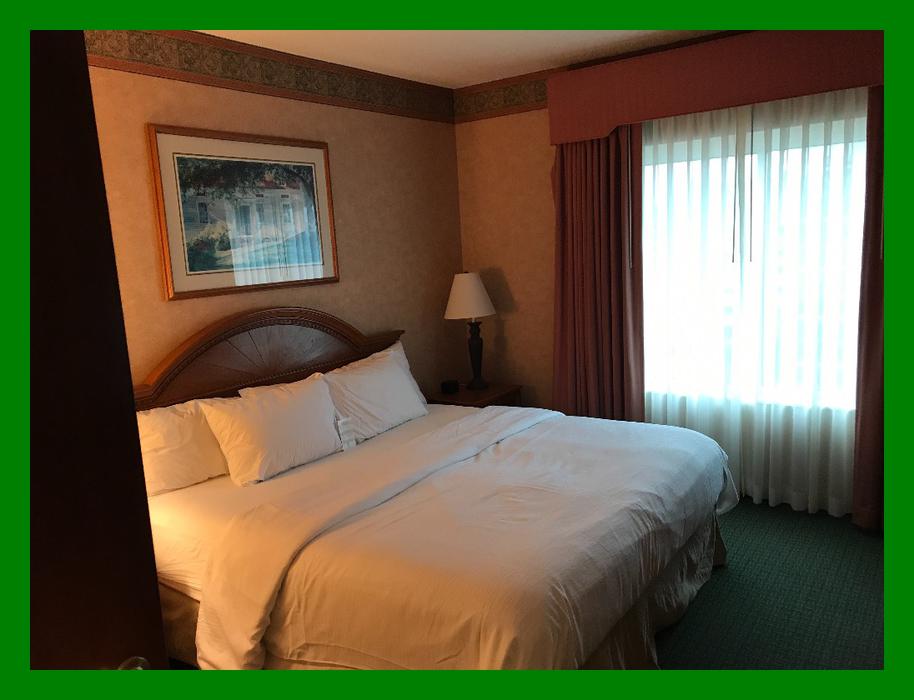} \\ \hline

   \multirow{4}{*}{\vspace{-1in}\includegraphics[height=1.75in]{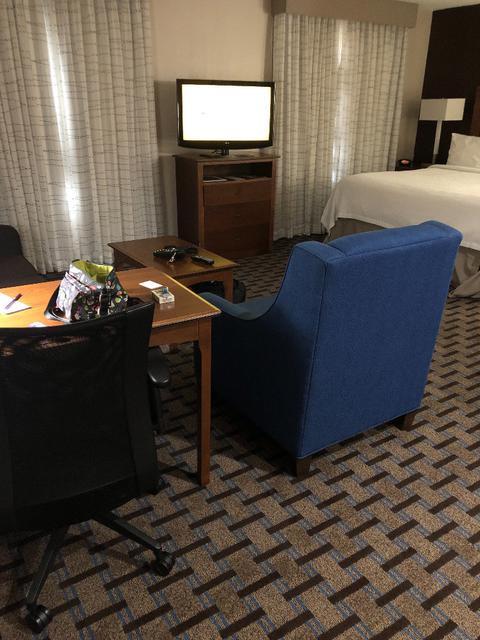}}  &
   \raisebox{2\height}{Cross-Entropy} & \includegraphics[height=\imheight]{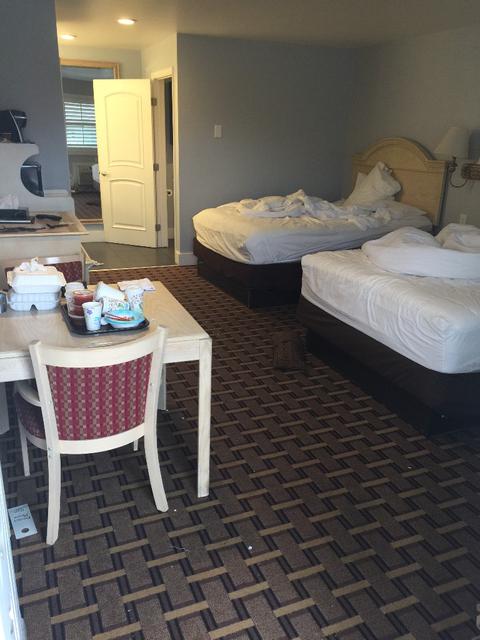} & \includegraphics[height=\imheight]{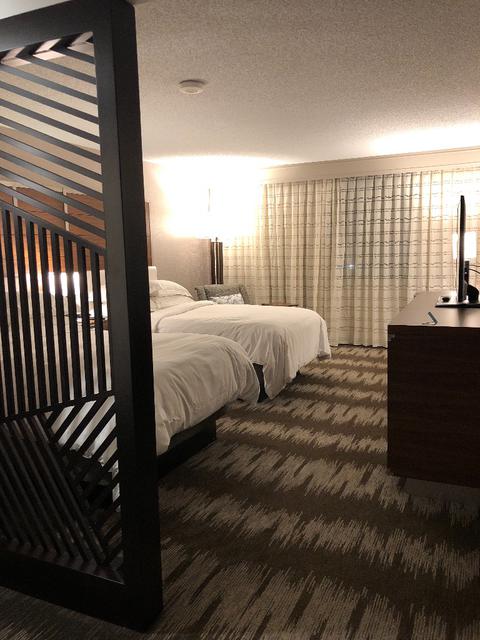} & \includegraphics[height=\imheight]{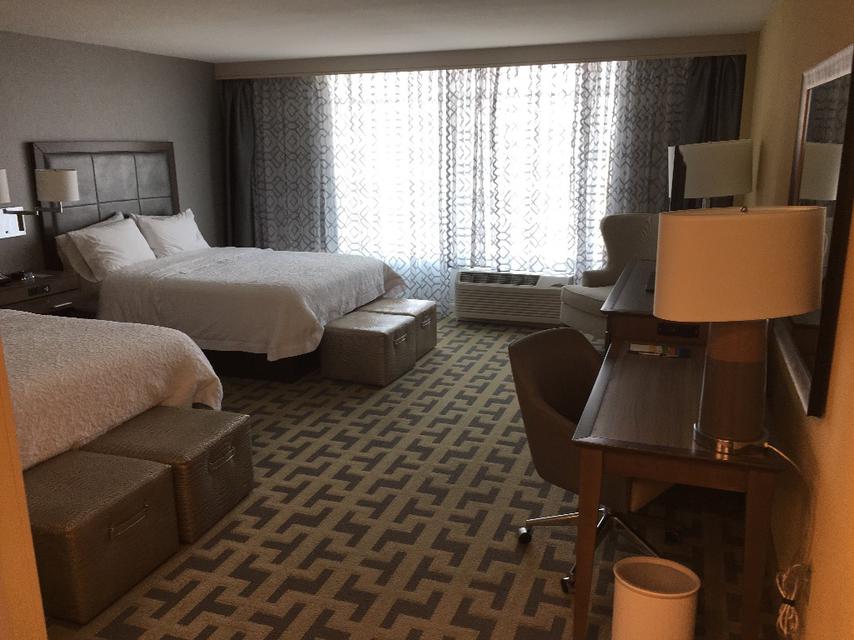} & \includegraphics[height=\imheight]{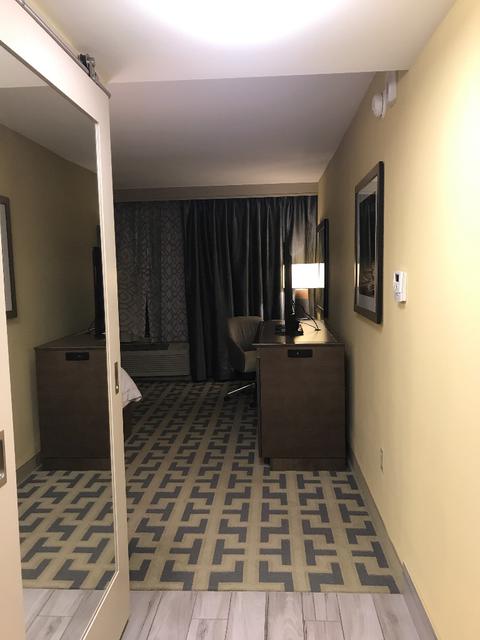} & \includegraphics[height=\imheight]{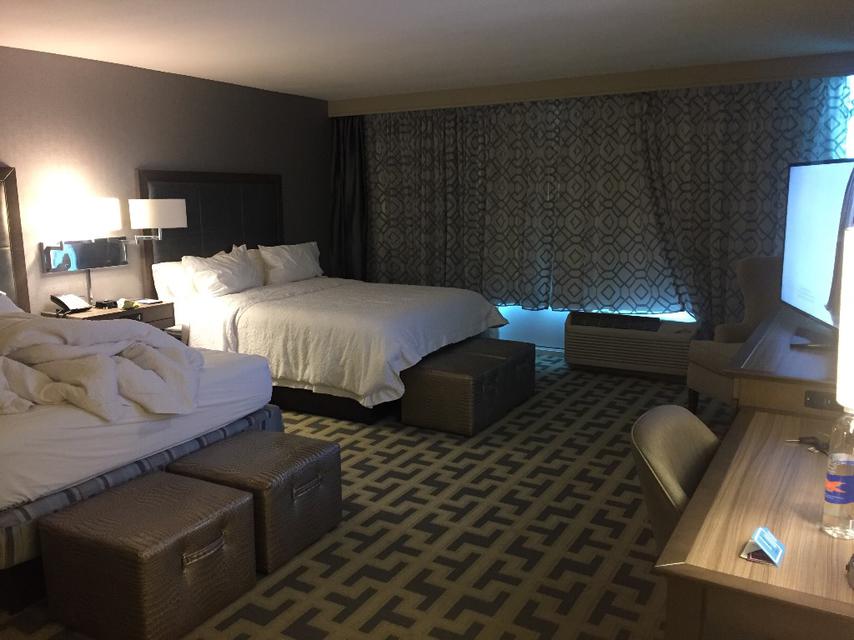} \\
 & \raisebox{2\height}{Batch-All} & \includegraphics[height=\imheight]{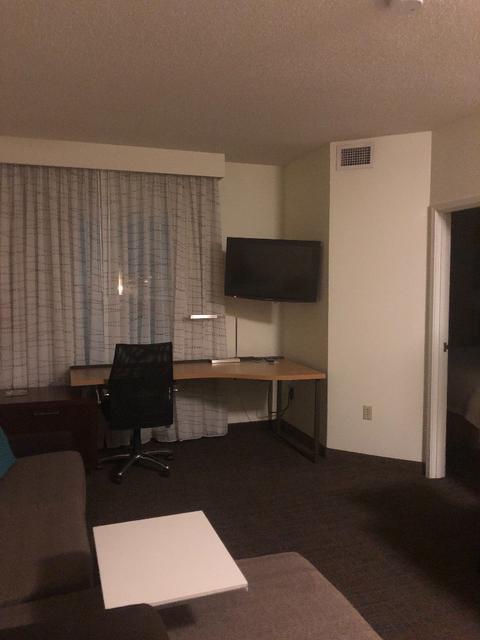} & \includegraphics[height=\imheight]{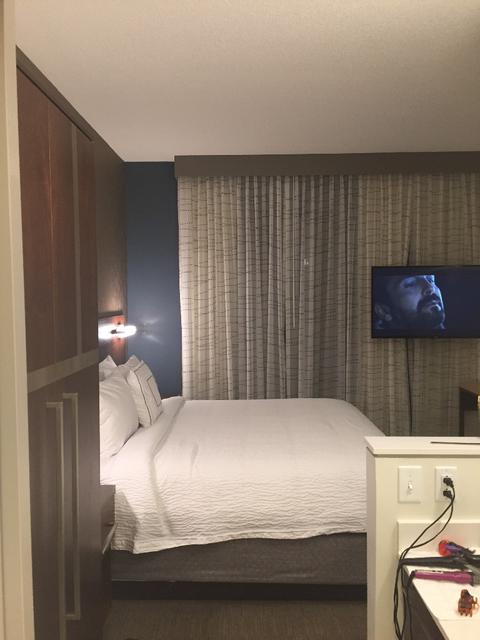} & \includegraphics[height=\imheight]{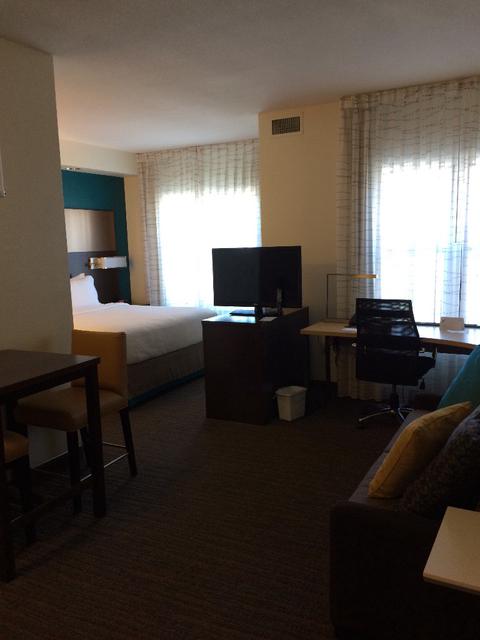} & \includegraphics[height=\imheight]{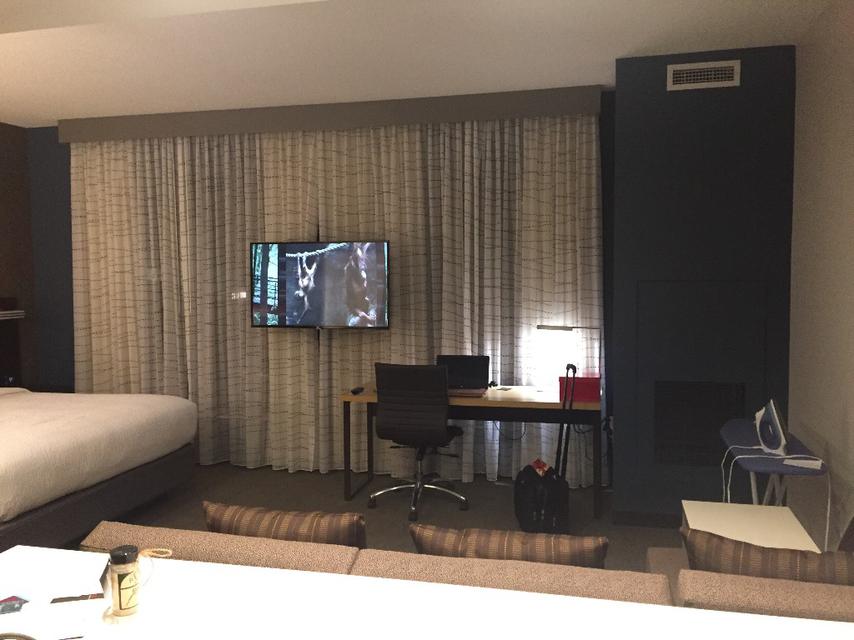} & \includegraphics[height=\imheight]{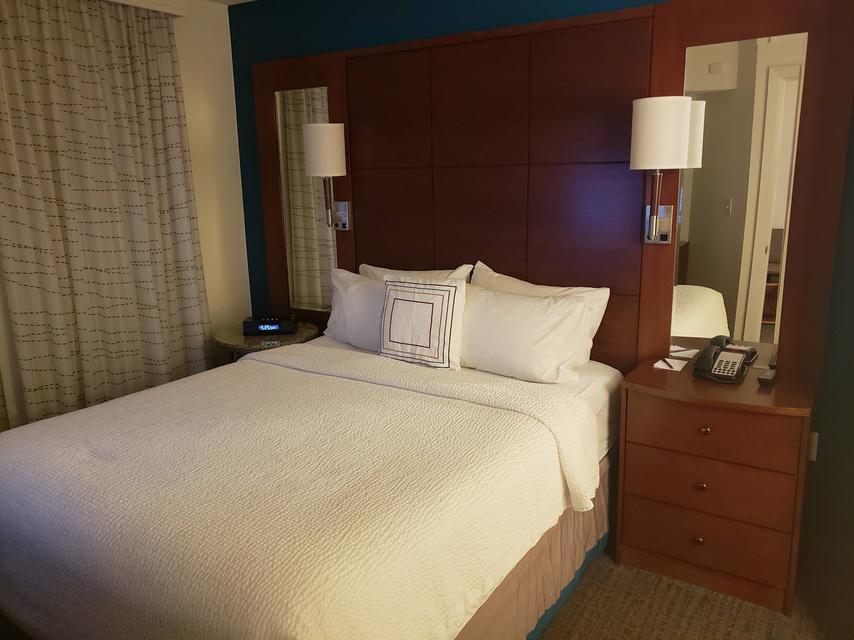} \\
  & \raisebox{2\height}{EPHN} & \includegraphics[height=\imheight]{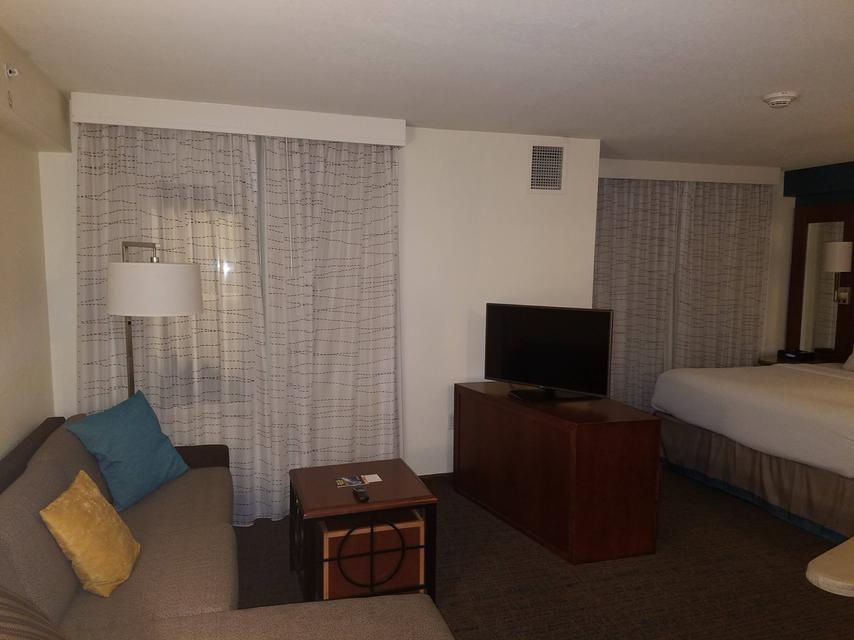} & \includegraphics[height=\imheight]{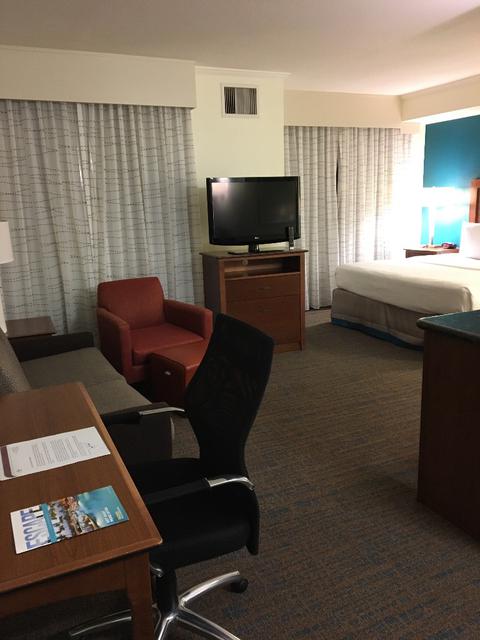} & \includegraphics[height=\imheight]{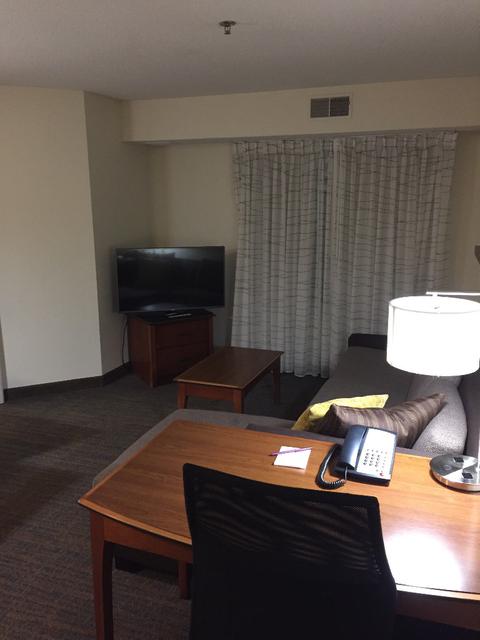} & \includegraphics[height=\imheight]{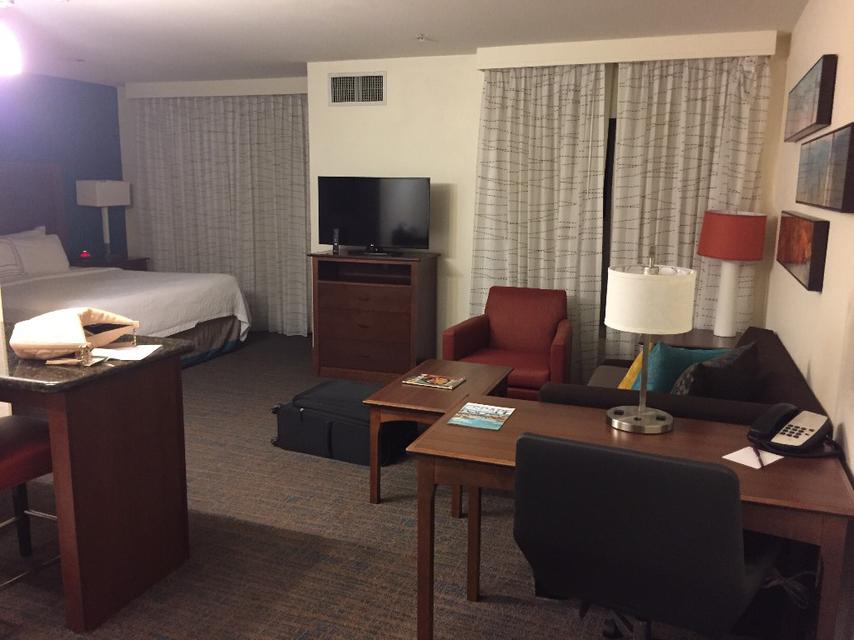} & \includegraphics[height=\imheight]{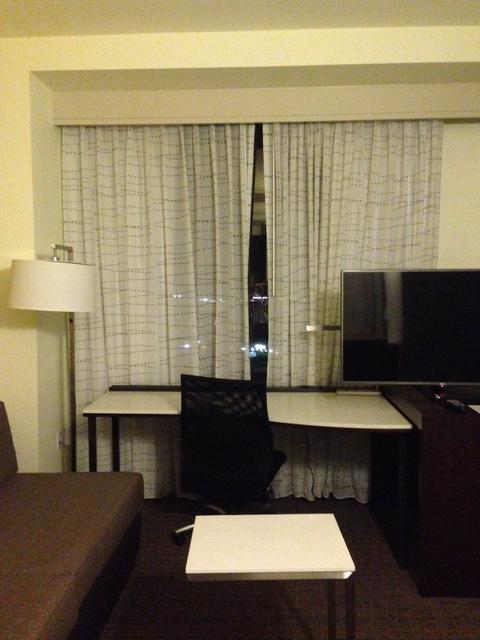} \\
   & \raisebox{2\height}{SCT} & \includegraphics[height=\imheight]{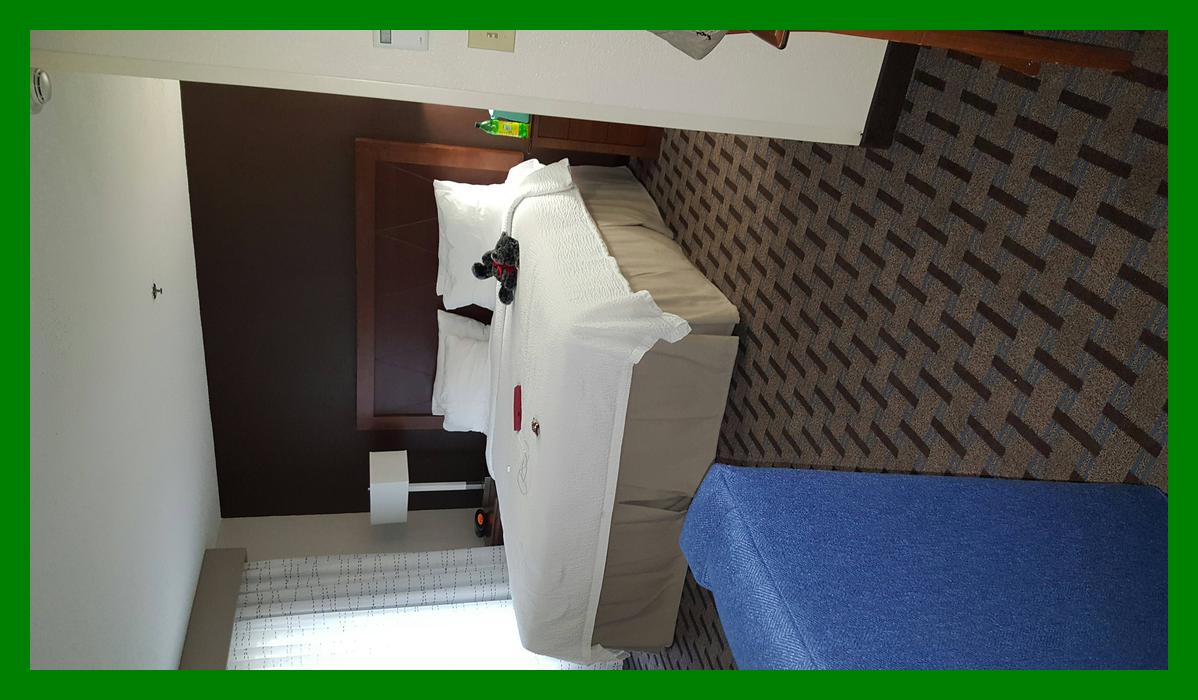} & \includegraphics[height=\imheight]{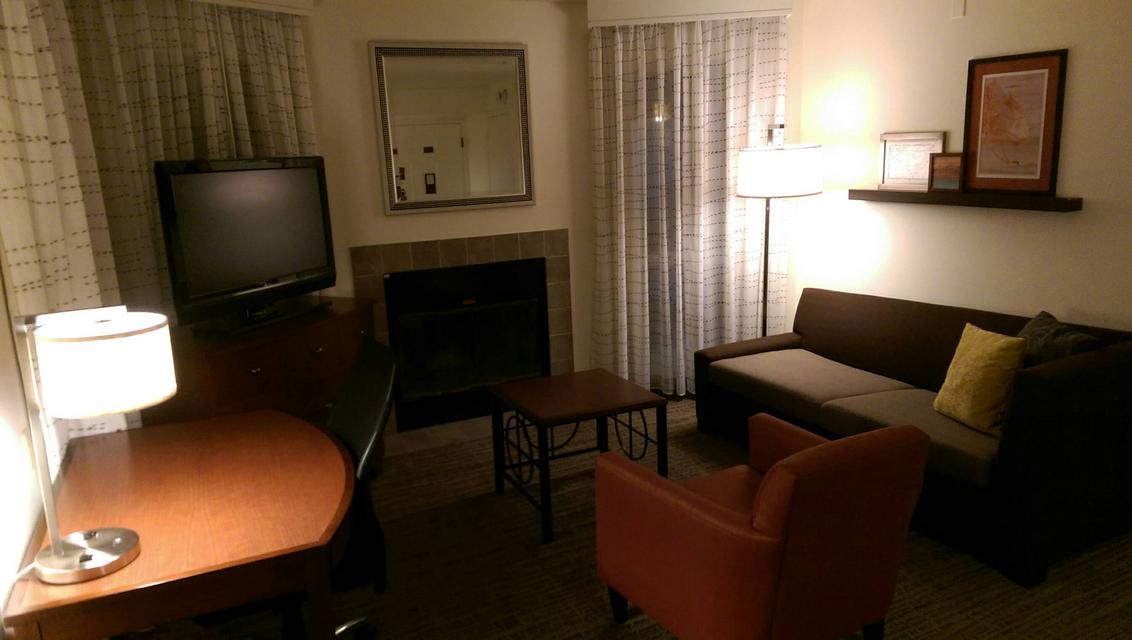} & \includegraphics[height=\imheight]{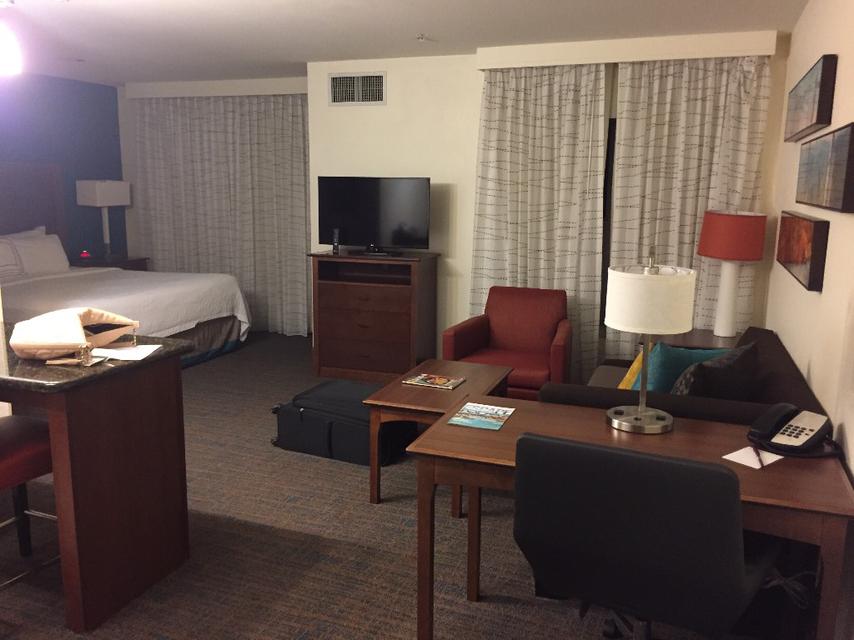} & \includegraphics[height=\imheight]{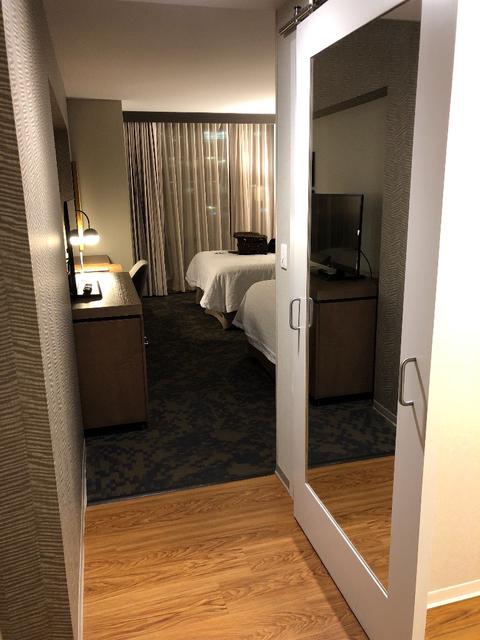} & \includegraphics[height=\imheight]{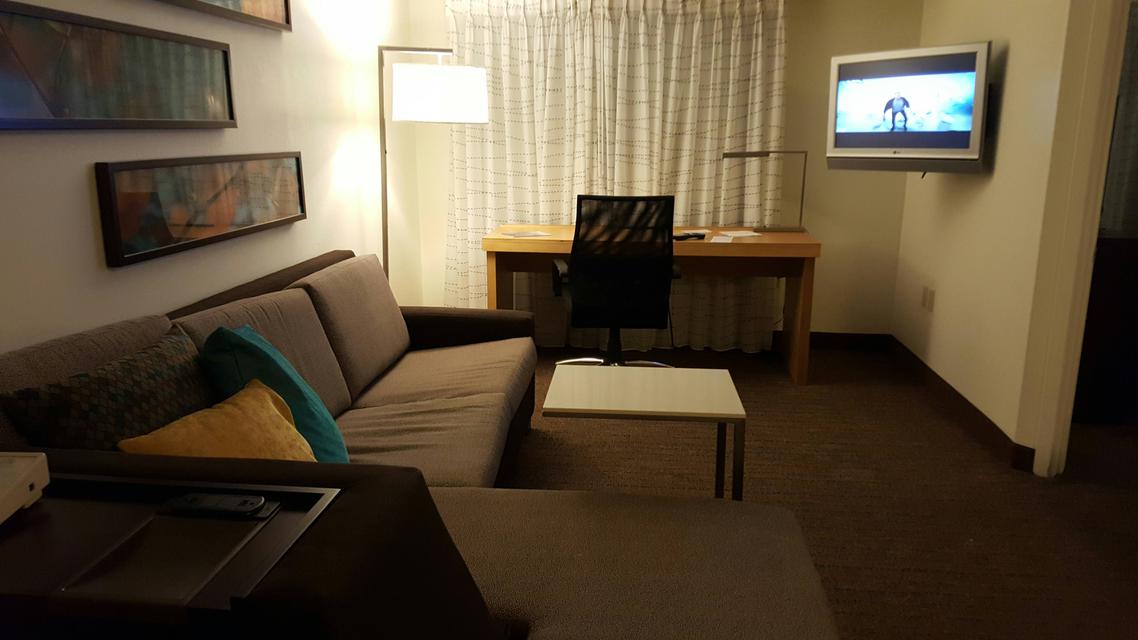} \\ \hline

   \multirow{4}{*}{\vspace{-1in}\includegraphics[height=1in]{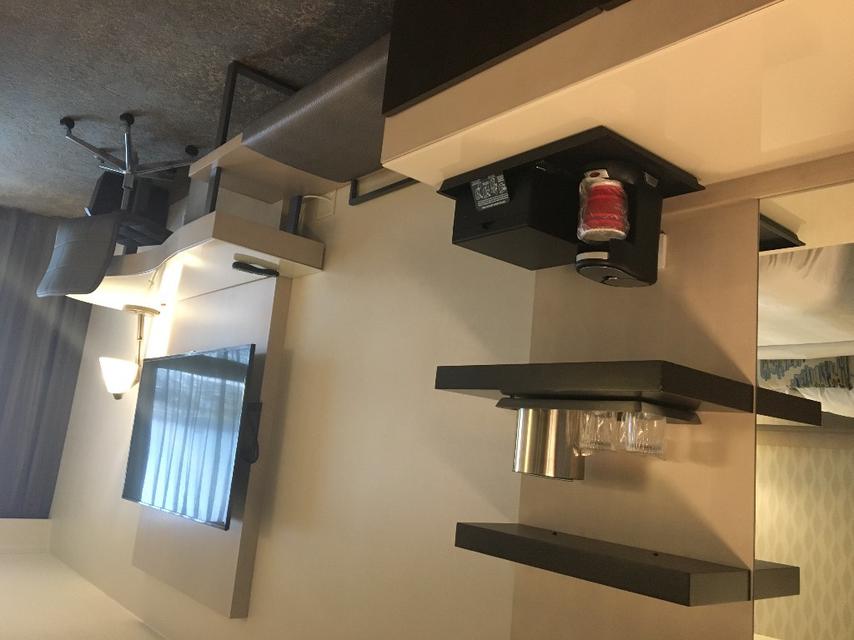}}  & \raisebox{2\height}{Cross-Entropy} & \includegraphics[height=\imheight]{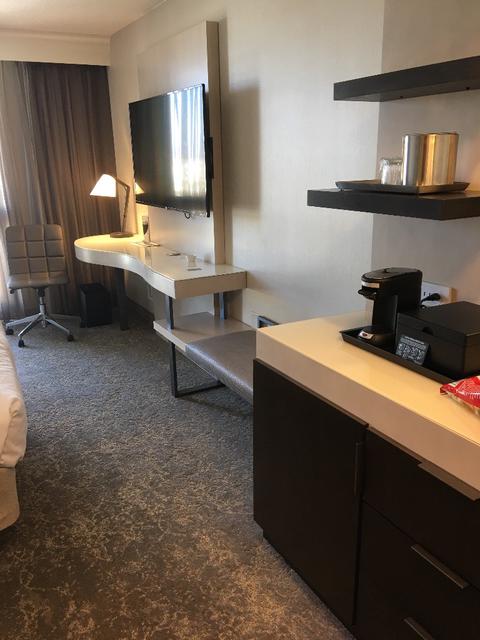} & \includegraphics[height=\imheight]{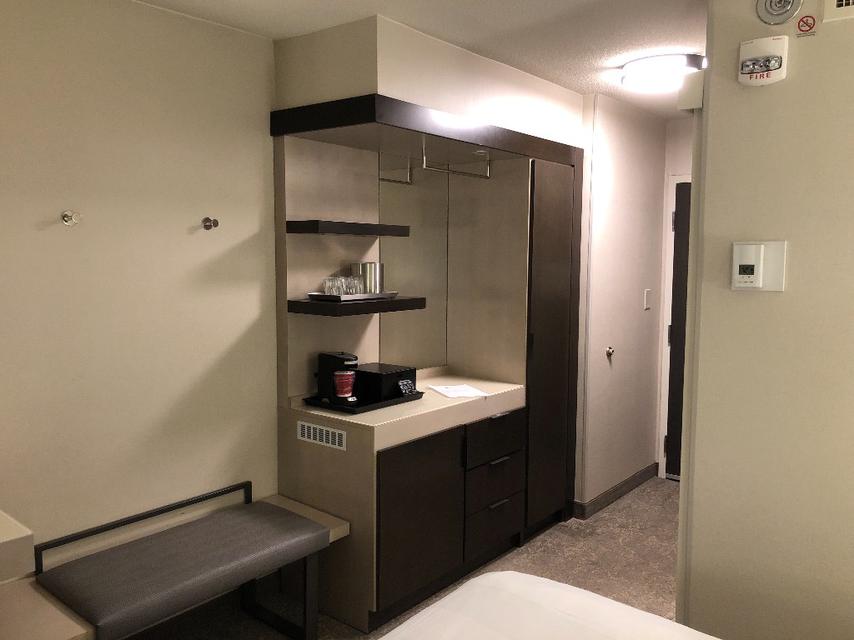} & \includegraphics[height=\imheight]{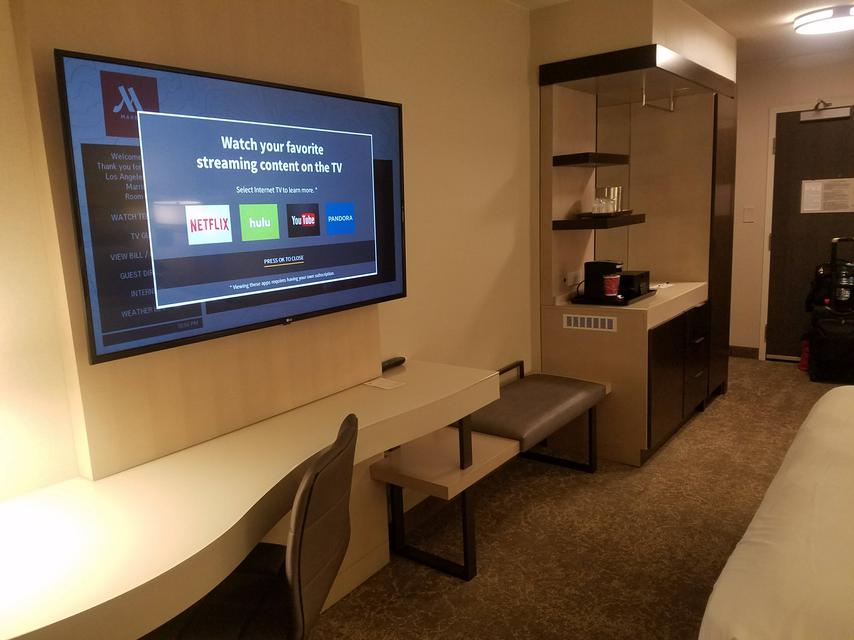} & \includegraphics[height=\imheight]{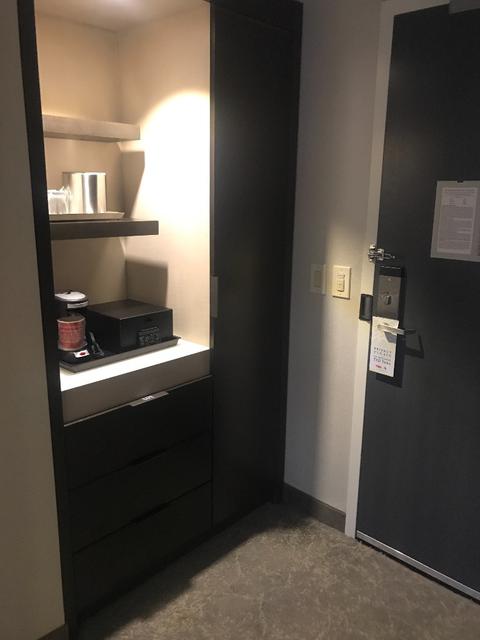} & \includegraphics[height=\imheight]{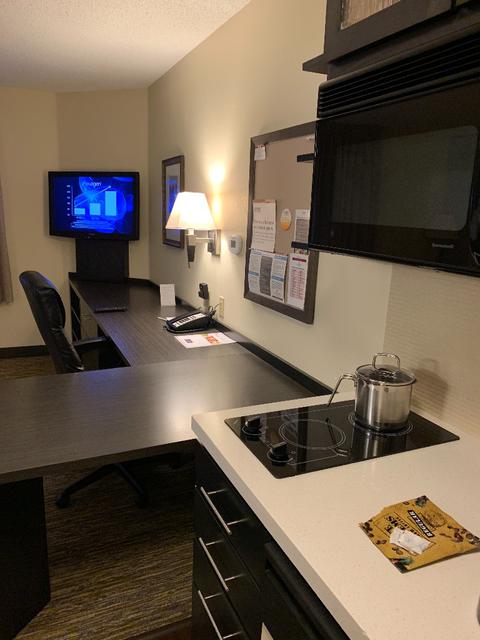} \\
 & \raisebox{2\height}{Batch-All} & \includegraphics[height=\imheight]{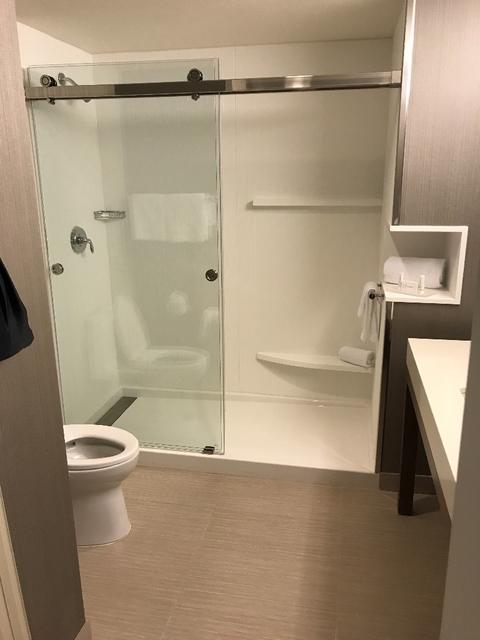} & \includegraphics[height=\imheight]{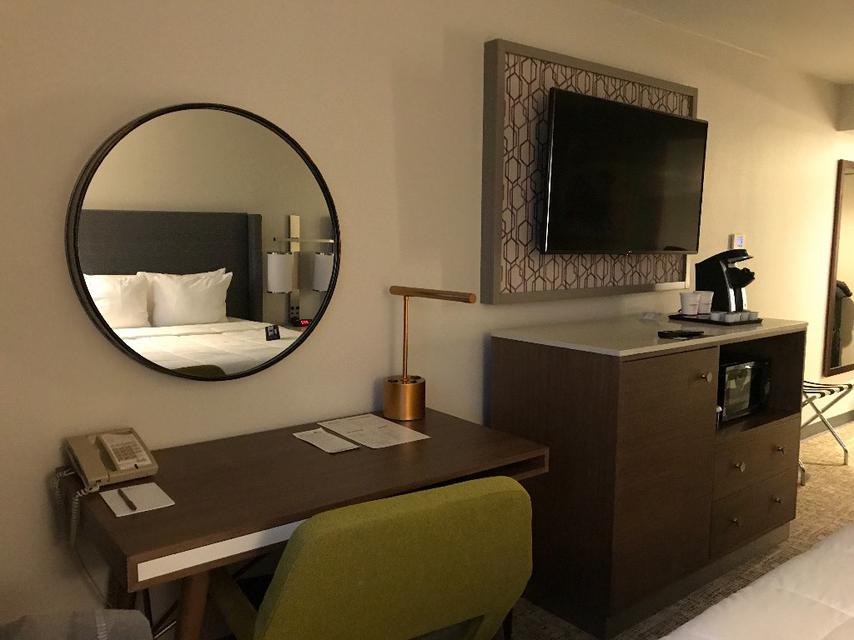} & \includegraphics[height=\imheight]{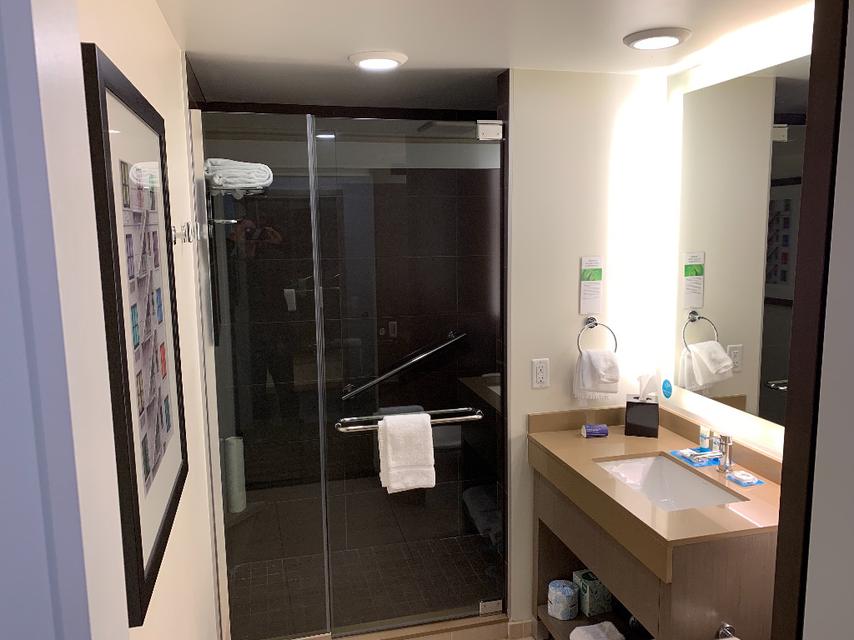} & \includegraphics[height=\imheight]{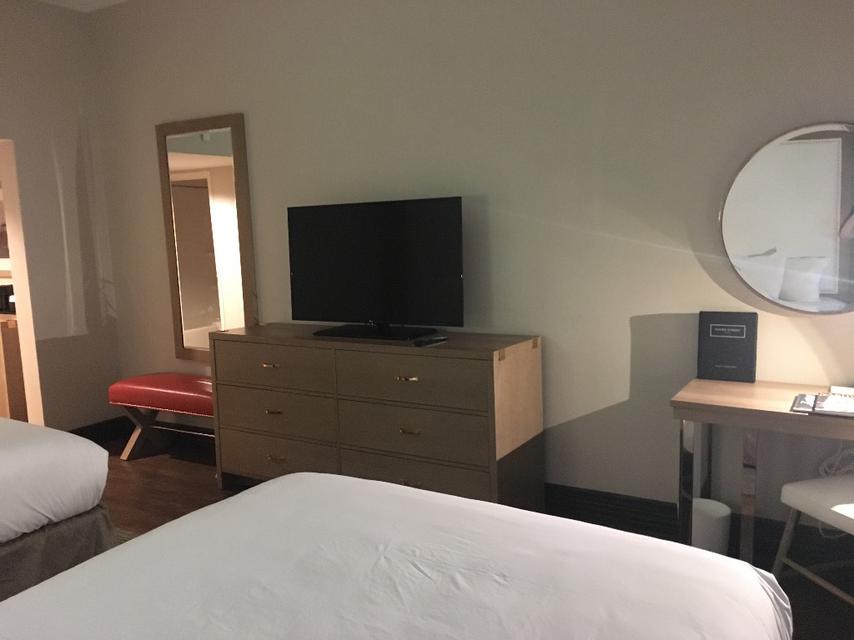} & \includegraphics[height=\imheight]{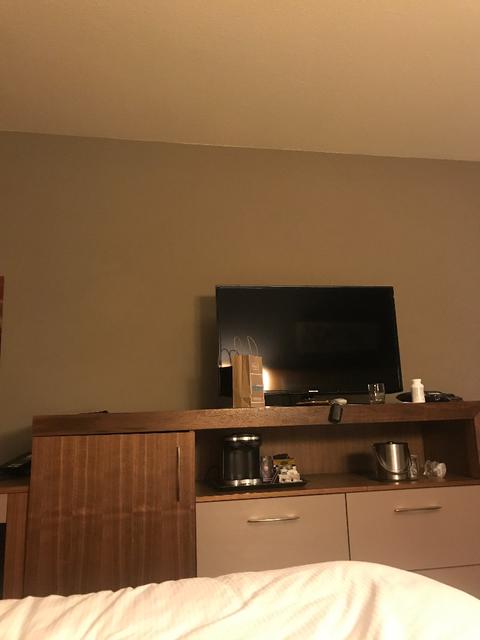} \\
  & \raisebox{2\height}{EPHN} & \includegraphics[height=\imheight]{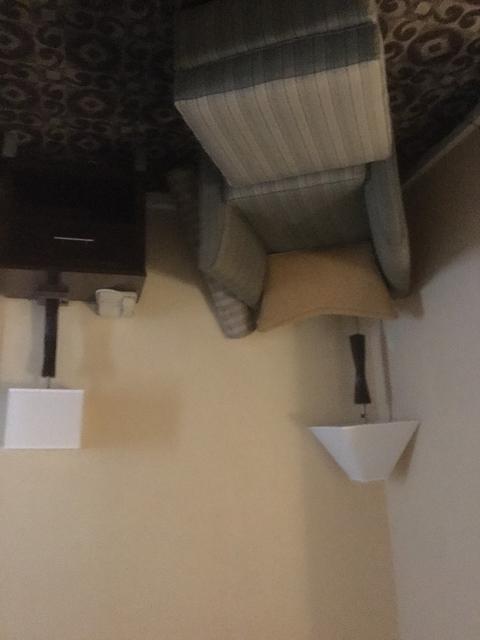} & \includegraphics[height=\imheight]{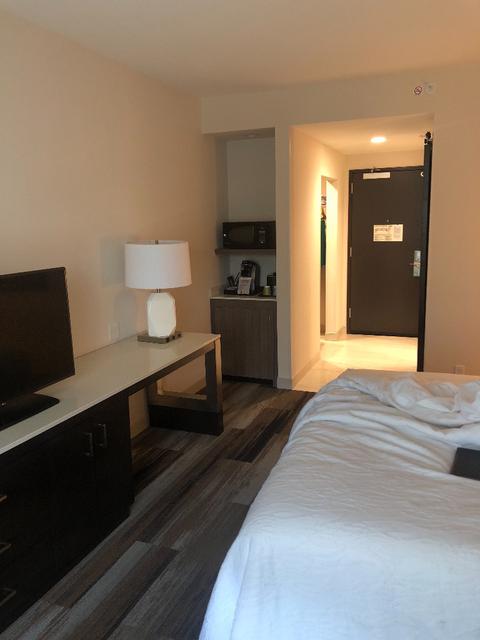} & \includegraphics[height=\imheight]{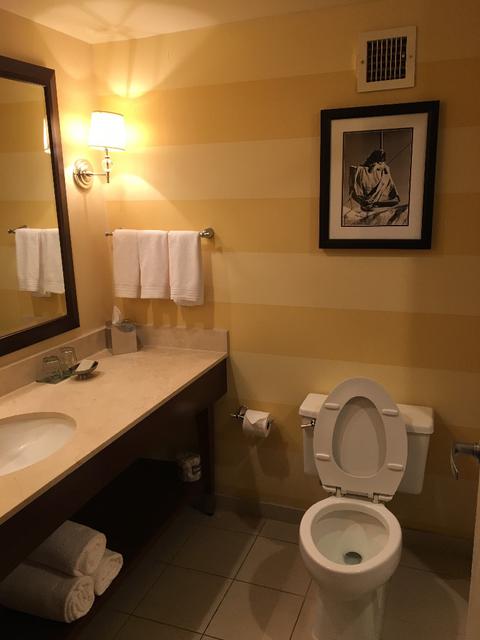} & \includegraphics[height=\imheight]{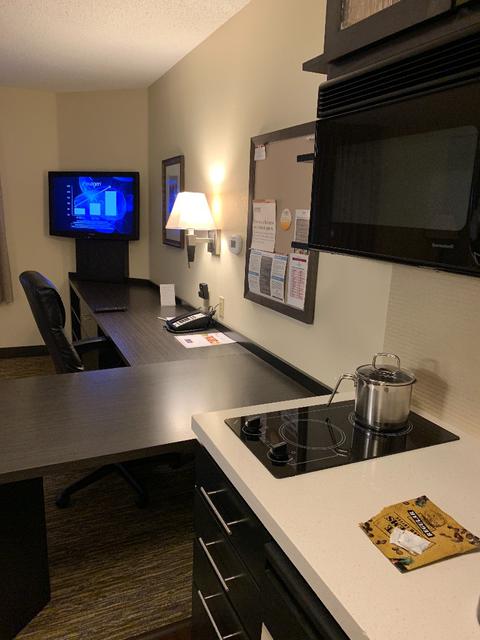} & \includegraphics[height=\imheight]{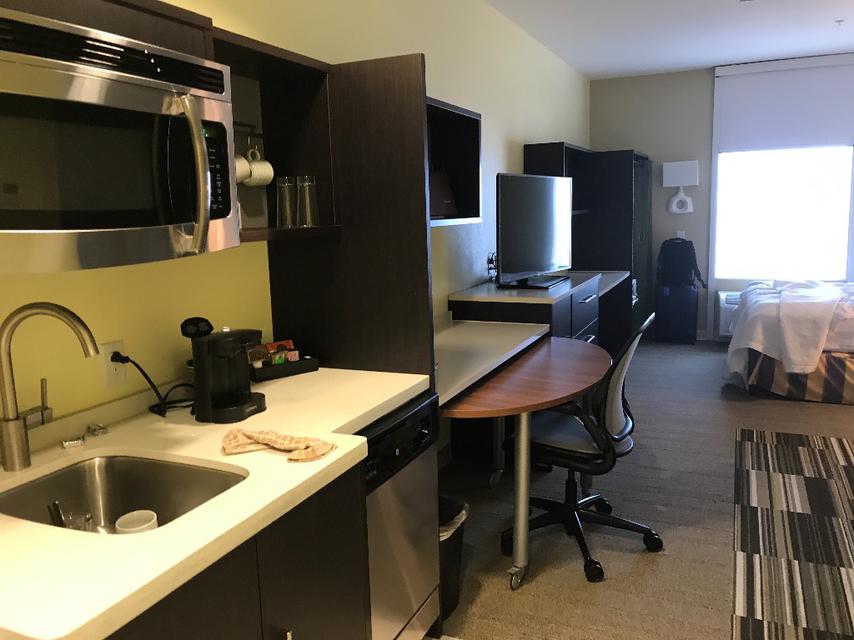} \\
   & \raisebox{2\height}{SCT} & \includegraphics[height=\imheight]{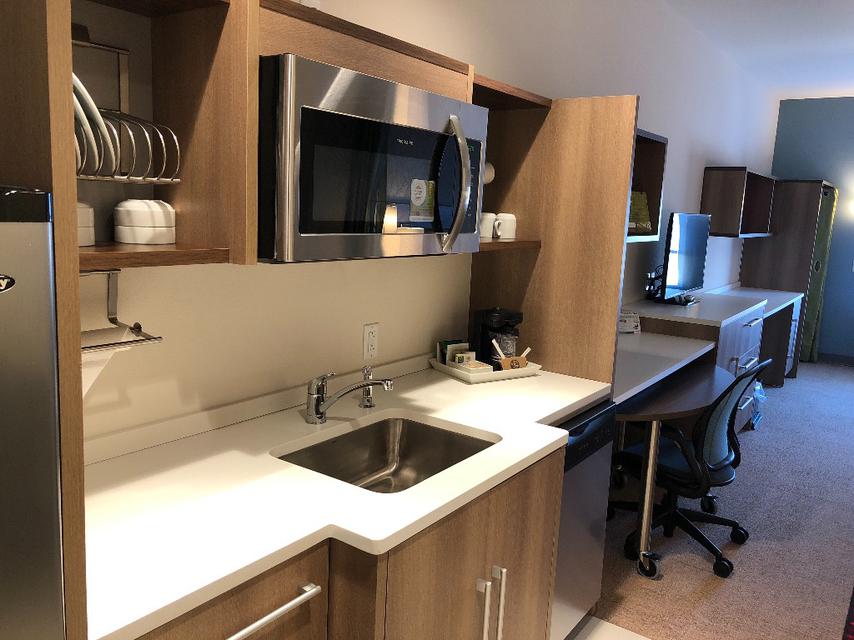} & \includegraphics[height=\imheight]{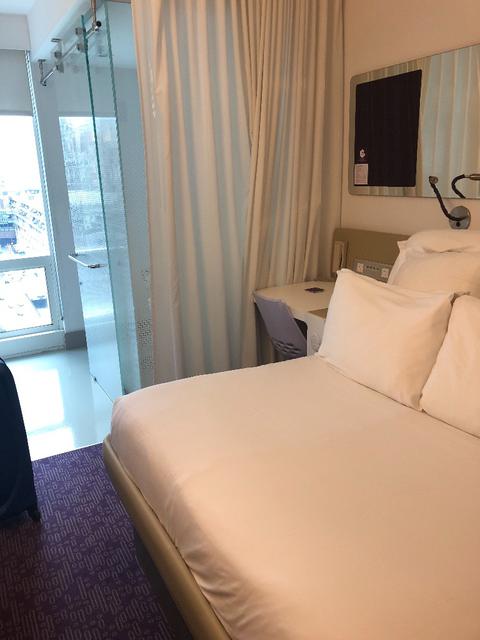} & \includegraphics[height=\imheight]{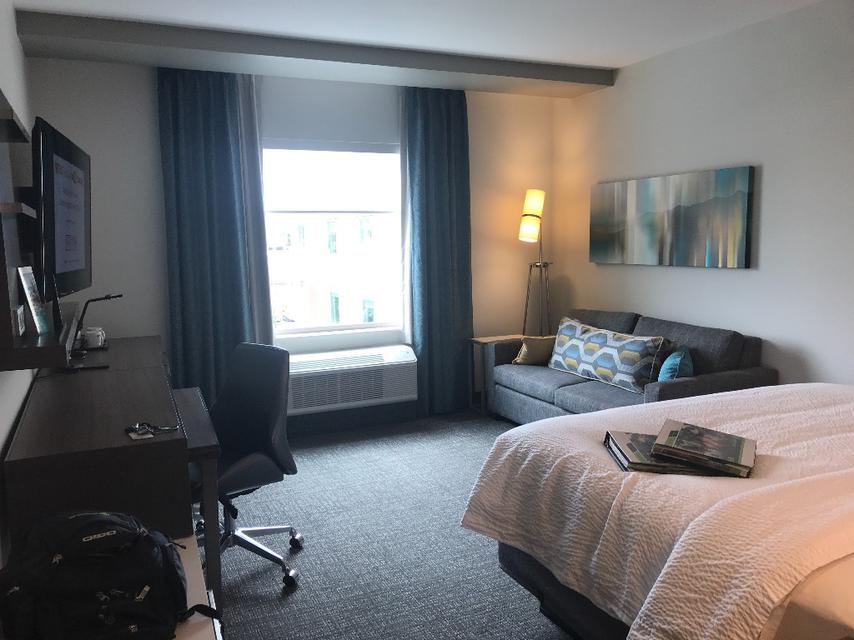} & \includegraphics[height=\imheight]{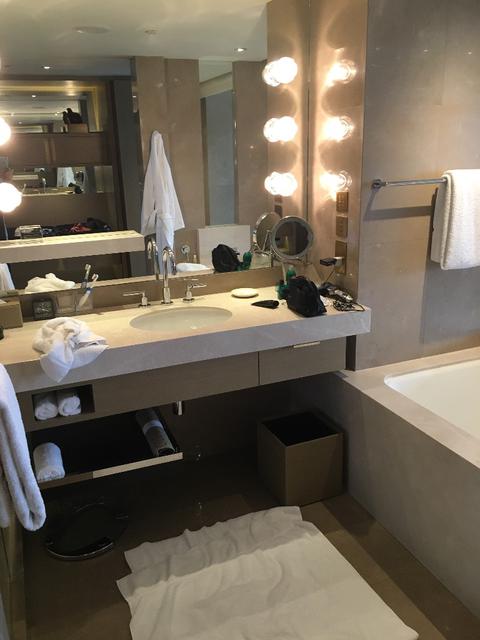} & \includegraphics[height=\imheight]{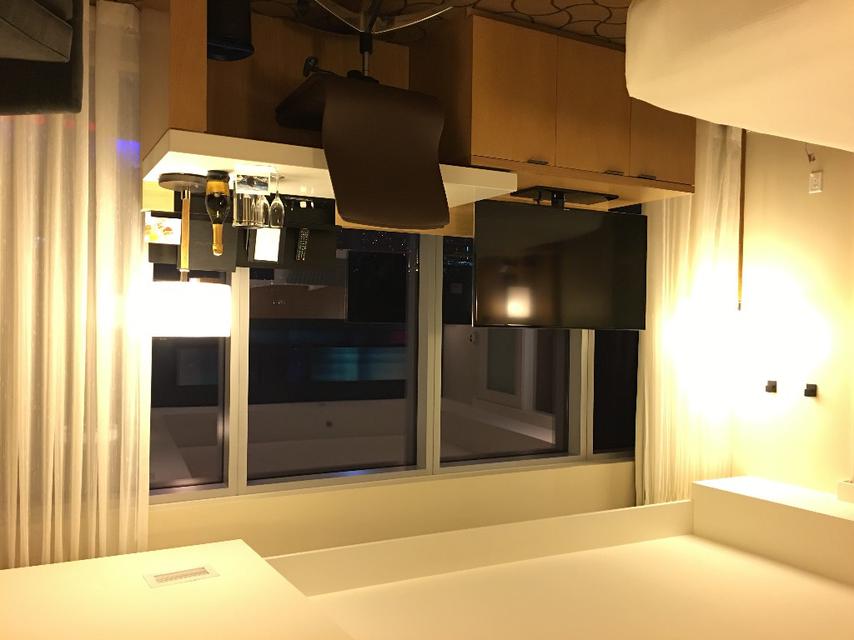} \\ \hline
\end{tabular}
\caption{Example queries from the Hotel-ID test set, shown with their 5 nearest neighbors (computed with cosine similarity) using each of the different evaluated approaches. Results from the same hotel as the query are outlined in green.}\label{fig:example_results}
\end{figure*}

\section{Ethical Concerns}\label{sec:ethical}
Human trafficking effects large numbers of innocent victims every year, with individuals being forced into labor or commercial sexual exploitation against their will for the profit of someone else~\cite{bouche2015report,nationalStrategy,ncmecAmicusBrief}. But the reality is that it's also an extremely complex issue and it can be hard to define whether an individual is being trafficked, versus using their own agency to pursue a job, as is often the case for sex workers who law enforcement all too often conflate with human trafficking victims. While the TraffickCam system, the Hotels-50K dataset, and this 2021 Hotel-ID dataset and challenge were created with the goal of helping to combat human trafficking, we cannot ignore that the same models that can recognize hotels in images of human trafficking victims could also be used by law enforcement to target sex workers, or be used for unrelated purposes, such as to locate undocumented immigrants photographed in hotels.

We have made concerted efforts to avoid such uses -- the image search tools that we have deployed have only been made available at the National Center for Missing and Exploited Children, an organization that only works on cases relating to minors, and even within that system we have built in monitoring tools to detect unusual uses or the possible sharing of credentials. We additionally have only ever publicly shared small enough subsets of data, in either this dataset or prior ones, to minimize the possibility of more nefarious actors being able to readily produce a similar system. Nonetheless, we recognize the inherent risk in working in this space, and encourage everyone working on this or related problems to consider those risks and how to take actions to mitigate them.

\section{Conclusion}
In this paper, we described the hotel recognition problem, how it is important in investigations of human trafficking. We presented the 2021 Hotel-ID dataset and included baseline results using both classification and deep metric learning. The dataset and challenge can be found at \url{https://www.kaggle.com/c/hotel-id-2021-fgvc8}.

{
\bibliographystyle{ieee_fullname}
\bibliography{cvpr}
}

\end{document}